\documentclass{article} 
\usepackage{iclr2026_conference,times}

\usepackage{amsmath,amsfonts,bm}

\def\eqref#1{equation~\ref{#1}}

\def\1{\bm{1}}

\DeclareMathAlphabet{\mathsfit}{\encodingdefault}{\sfdefault}{m}{sl}
\SetMathAlphabet{\mathsfit}{bold}{\encodingdefault}{\sfdefault}{bx}{n}

\usepackage[hypertexnames=false]{hyperref}
\usepackage{url}
\usepackage{graphicx}
\usepackage{amsthm}
\newtheorem{assumption}{Assumption}
\newtheorem{theorem}{Theorem}
\graphicspath{{figures/}}
\usepackage{algorithm}
\usepackage{algorithmic}

\title{Backpropagation-Free Trunk Training via the Split Forward Gradients}

\author{Tian Qin \\
Math department\\
Lehigh University\\
Bethlehem, PA 18015, USA \\
\texttt{\{tiq218\}@lehigh.edu} \\
\And
Wei-Min Huang \\
Math department \\
Lehigh University\\
Bethlehem, PA 18015, USA \\
\texttt{\{wh02\}@lehigh.edu} \\
}

\iclrfinalcopy 
\begin{document}

\maketitle
\lhead{Preprint}
\begin{abstract}
Backpropagation makes training deep networks memory intensive because it must
store intermediate activations. Forward-mode methods avoid this cost, but
their gradient estimates become increasingly noisy as the number of trained
parameters grows. We introduce \emph{Split Forward Gradient} (Split-FG), which
splits a network at an intermediate representation: it computes the output
head gradient exactly and estimates only the trunk gradient with a
Jacobian--vector product. This reduces estimator variance and requires no
backward pass through the trunk, while retaining an Adam-style convergence
guarantee. Our experiments reveal an important practical failure mode. On WikiText-103,
naive forward-gradient training of the trunk performs worse than leaving a
randomly initialized trunk frozen, likely because Adam updates every noisy,
under-determined trunk coordinate too aggressively. Simply using a much
smaller learning rate for the trunk reverses this result: a $16$M-parameter
GPT-2-style model reaches validation perplexity $387$, compared with $668$ for
the frozen-trunk control and $2{,}885$ for a matched pure forward-gradient
baseline (backpropagation reaches $150$). Split-FG also produces the strongest
backprop-free results on our tabular benchmarks and reaches $60.5\%$ on
CIFAR-10 and $35.2\%$ on CIFAR-100 with a heavy-head design. It reduces peak
memory by up to $35\%$ relative to matched backpropagation, although the
performance gap widens as the forward-mode trunk grows.
\end{abstract}

\section{Introduction}

The computational cost of training deep neural networks is dominated by backpropagation \citep{Rumelhart1986LearningRB}: the reverse-mode differentiation of a loss through the entire network via the chain rule. This requires storing the full computational graph and all intermediate activations in memory—a cost that scales linearly with network depth and sequence length: at roughly $16\,d_{\mathrm{model}}$ bytes per token per layer (half precision), a $32$-layer, $d_{\mathrm{model}}{=}4096$ model at sequence length $4096$ and batch size $8$ holds ${\approx}69$\,GB of activations alone, beyond a single $80$\,GB accelerator once weights and optimizer state are added. These requirements are obstacles for deployment on resource-constrained hardware, for continual learning systems that update parameters online, and for models of biological learning, where symmetric backward error transport through deep stacks is considered implausible (though our own exact head still uses weight transport at its single readout layer). A sharper case than cost is hardware where the backward pass is not merely expensive but unavailable: optical and analog accelerators run the forward pass natively at very low energy \citep{Shen2016DeepLW}, but backpropagating through them requires the adjoint method---reverse field propagation with in-situ measurement, or an accurate digital twin \citep{hughes2018training}. What such hardware admits is gradient information from forward evaluations alone, directional derivatives obtained by perturb-and-measure, with no reverse pass through the physical stack-, which is precisely the regime the method developed for backpropagation-free targets.

Prior work on backpropagation-free training has demonstrated promising results on small-scale tasks. Forward gradients via JVP \citep{baydin2022gradientsbackpropagation} and evolution strategies \citep{salimans2017evolutionstrategiesscalablealternative,JMLR:v15:wierstra14a} provide unbiased gradient estimates but suffer from variance that scales linearly with parameter count. Zeroth-order methods such as MeZO \citep{malladi2023finetuning} can fine-tune pretrained language models but struggle with pretraining from scratch. Equilibrium propagation \citep{EP,EPL}  and predictive coding \citep{Whittington2017AnAO,millidge2022predictive,Salvatori2021ReverseDV} match backpropagation on shallow networks through local energy minimisation, but require iterative settling phases that multiply per-step cost. The Forward-Forward algorithm \citep{hinton2022forwardforwardalgorithmpreliminaryinvestigations} replaces the backward pass with contrastive positive/negative phases per layer, though it has only been validated on small classifiers. Direct feedback alignment \citep{Nkland2016DirectFA} and local learning rules \citep{ren2023scaling} reduce inter-layer dependence but introduce bias through fixed random projections or surrogate losses. Despite this breadth, a persistent gap remains: these methods degrade when scaled to transformers, need a pretrained initialisation, or introduce systematic bias through approximate objectives.

In this paper we show that decomposing the gradient computation at an
intermediate representation substantially reduces forward-gradient variance
and recovers part of the gap to backpropagation, with no backward pass through
the trunk. The strict, unclipped estimator is unbiased; our runs apply global
gradient clipping, and the GPT runs additionally train the tied vocabulary
table through a readout-only surrogate (Appendices~\ref{app:general-protocol}
and~\ref{app:gpt-arch}). We call the estimator \emph{Split
Forward Gradient} (Split-FG) and make the following contributions.
\begin{enumerate}

\item \textbf{The Split-FG estimator.} We decompose the forward gradient at
any intermediate representation~$h$, compute the head-side gradient
$\partial\mathcal{L}/\partial h$ in closed form (or estimate it in low
dimension), and restrict the stochastic forward gradient to the trunk. The
directional derivative is exact with respect to the head, so only the trunk
contributes noise. We evaluate the Split-FG estimator on tabular, image, and language
tasks and shows its superiority over many backpropagation free baselines.

\item \textbf{A measured variance analysis.} We show that the  Split-FG estimator enables per-coordinate estimator
variance to drop from $\lVert\nabla_{\theta}\mathcal{L}\rVert^2$ to
$\lVert\nabla_{\theta_{\mathrm{trunk}}}\mathcal{L}\rVert^2$, and we measure
this factor rather than assuming it. On our GPT configuration, where the
vocabulary readout holds $80\%$ of the parameters, the trunk carries $0.74$
of the gradient energy at initialisation, giving a $1.35\times$ per-coordinate
and $6.8\times$ total variance reduction (Section~\ref{split-fg}). The exact
head gradients are available in closed form for linear cross-entropy and MSE
heads, and for a factored two-linear head, so the pipeline uses no
reverse-mode automatic differentiation.

\item \textbf{An empirical failure mode and its fix.} Using a frozen-trunk
control that trains only the exact head over a fixed random trunk, we find
that naive forward-gradient training of the trunk is worse than not training
it at all ($1733$ versus $668$ perplexity on WikiText-103). We attribute this
to Adam, which steps every noisy, under-determined trunk coordinate at full
length. Scaling the trunk step down by a factor $\rho=0.03$ removes the
failure: a $16$M-parameter GPT-2-style transformer trained from scratch
reaches perplexity $387$, which is $42\%$ below the frozen control and
$7.5\times$ below a pure forward-gradient baseline matched in tangent count
and tuning ($2{,}885$ at $K{=}4$). In these validation runs it uses
$35\%$ less peak memory than the same-architecture Adam backpropagation
reference (which reaches $150$), but takes $3.2\times$ longer per step.

\end{enumerate}

The remainder of the paper is organized as follows. Section~\ref{preliminary} develops Split-FG and its variance analysis; Section~\ref{sec:theory} gives a convergence guarantee for a predictable-preconditioner idealisation of the split estimator. Section~\ref{sec:experiments} compares Split-FG with backpropagation and no-trunk-backward baselines across tabular, image, and GPT-style language-modelling tasks, where a frozen-trunk control isolates the trunk's contribution and motivates the trunk-step fix (Section~\ref{sec:main-experiment}; Appendix~\ref{sec:frozen-control}), followed by ablation studies. Section~\ref{discussion} discusses limitations and future directions. Code for all experiments is provided in the supplementary material.

\section{Preliminaries}
\label{preliminary}

\subsection{Forward Gradient}

Consider a neural network parameterised by 
$\theta \in \mathbb{R}^P$ with scalar loss 
$\mathcal{L}(\theta)$. For a direction vector 
$v \in \mathbb{R}^P$, the directional derivative
is the scalar
\begin{equation}
\label{eq:dir-deriv}
d \;=\; \nabla_{\theta}\mathcal{L}(\theta) \cdot v 
  \;=\; \sum_{j=1}^{P} 
        \frac{\partial \mathcal{L}}{\partial \theta_j}\, v_j\,.
\end{equation}
This can be computed exactly, without a backward pass, via a 
Jacobian-vector product (JVP) using forward-mode 
automatic differentiation 
\citep{griewank2008evaluating}. The JVP augments each
intermediate variable with a tangent and propagates both jointly through the
computation graph via dual-number arithmetic; initialising the tangent of
$\theta$ to $v$ yields the pair $(\mathcal{L},\, d)$ in a single pass.
The computational cost is comparable to one forward evaluation; 
crucially, no activations need to be stored for a backward pass. Writing
$g := \nabla_{\theta}\mathcal{L}(\theta)$ with coordinates $g_i$,
\citet{baydin2022gradientsbackpropagation} observe that if
$v \sim \mathcal{N}(0, I_P)$, the product $d \cdot v$ is an
\textbf{unbiased estimate} of the full gradient:
\begin{equation}
\label{eq:fg-estimator}
\hat{g} = d \cdot v\,, \qquad
\mathbb{E}[\hat{g}] = g\,.
\end{equation}
Unbiasedness is immediate from $\mathbb{E}[v\,v^\top] = I$; the
per-coordinate computation appears in Appendix~\ref{app:variance-proof}.
With $K$ independent samples
$\{v_k\}_{k=1}^{K}$, the averaged estimator
\begin{equation}
\label{eq:fg-K}
\hat{g}^{(K)}
= \frac{1}{K} \sum_{k=1}^{K} d_k\, v_k
\end{equation}
retains unbiasedness. Its per-coordinate variance follows from Isserlis'
theorem \citep{isserlis1918formula} on fourth-order Gaussian moments: since
$\mathbb{E}[v_j\,v_k\,v_i\,v_i]
= \delta_{jk} + 2\,\delta_{ji}\,\delta_{ki}$,
we obtain
\begin{equation}
\label{eq:fg-variance}
\mathrm{Var}\bigl[\hat{g}^{(K)}_i\bigr]
= \frac{1}{K}\Bigl(
  \lVert g \rVert^2 + g_i^2
  \Bigr)\,.
\end{equation}
The dominant term
$\lVert g\rVert^2
= \sum_{j=1}^{P} g_j^2$
sums over \emph{all} $P$ parameters, yielding a
signal-to-noise ratio for $i$-th coordinate:
\begin{equation}
\label{eq:snr}
\mathrm{SNR}_i
= \frac{g_i^2}
       {\lVert g\rVert^2}
\;\approx\; \frac{1}{P}\,.
\end{equation}
For a model with $P = 10^7$ parameters, each coordinate's 
signal is overwhelmed by noise from the remaining 
$P - 1$ coordinates. This linear scaling of variance 
with parameter count is the fundamental barrier to applying 
forward gradients at scale, and motivates the decomposition we 
introduce in Section~\ref{split-fg}.

\subsection{Split Forward Gradient}
\label{split-fg}
The core identity underlying Split-FG is easiest to read as an algorithmic
``push--score--estimate'' chain rule at an intermediate representation
$h = f(x;\,\theta_{\mathrm{trunk}}) \in \mathbb{R}^{d_h}$. Algorithm~\ref{alg:split-fg}
spells out the estimator. For each random direction in trunk-parameter space,
we first push the direction to the representation space, score the resulting
representation perturbation by the loss gradient at $h$, and then place that
scalar back on the same random direction.

\begin{algorithm}[t]
\caption{Split-FG estimator at hidden state \(h\)}
\label{alg:split-fg}
\begin{algorithmic}[1]
\REQUIRE Batch \(x,y\); split parameters
\(\theta=(\theta_{\mathrm{trunk}},\theta_{\mathrm{head}})\); sample count \(K\)
\ENSURE Gradient estimate \(\widehat g_{\mathrm{split}}^{(K)}\)
\STATE Compute the trunk representation
       \(h=f(x;\theta_{\mathrm{trunk}})\)
\STATE Compute the head loss and exact head-side quantities:
       \[
       \widehat g_{\mathrm{head}}
       =
       \nabla_{\theta_{\mathrm{head}}}\mathcal{L},
       \qquad
       u
       =
       \frac{\partial \mathcal{L}}{\partial h}
       \]
\STATE Let
       \(J_{\mathrm{trunk}}
       =
       \partial h/\partial\theta_{\mathrm{trunk}}\)
\FOR{\(k=1,\ldots,K\)}
    \STATE Sample
           \(v_k\sim\mathcal{N}(0,I_{P_{\mathrm{trunk}}})\)
    \STATE Compute the trunk JVP
           \(\Delta h_k=J_{\mathrm{trunk}}v_k\)
    \STATE Score the representation perturbation:
           \[
           d_k
           =
           \langle u,\Delta h_k\rangle
           =
           \left\langle
           J_{\mathrm{trunk}}^{\top}u,\,
           v_k
           \right\rangle
           =
           \left\langle
           \nabla_{\theta_{\mathrm{trunk}}}\mathcal{L},\,
           v_k
           \right\rangle
           \]
\ENDFOR
\STATE Average the forward-gradient samples on the trunk:
       \[
       \widehat g_{\mathrm{trunk}}^{(K)}
       =
       \frac{1}{K}\sum_{k=1}^{K}d_k\,v_k
       \]
\STATE Return
       \(\widehat g_{\mathrm{split}}^{(K)}
       =
       \widehat g_{\mathrm{trunk}}^{(K)}
       \oplus
       \widehat g_{\mathrm{head}}\),
       where \(\oplus\) concatenates the trunk and head estimates
\end{algorithmic}
\end{algorithm}

The standard forward gradient draws
\(v\in\mathbb{R}^{P_{\mathrm{total}}}\) and estimates all parameters from one
noisy scalar; Split-FG restricts the random directions to the trunk subspace and
computes the head gradient exactly. The chain-rule equality in the score step
of Algorithm~\ref{alg:split-fg} is the justification only: the implementation
computes each $d_k$ from $u = \partial\mathcal{L}/\partial h$ and $\Delta h_k$,
so no backward pass through the trunk ever occurs.
Appendix \ref{app:variance-proof} shows the estimator in
Algorithm~\ref{alg:split-fg} yields
a per-coordinate variance reduction of
%
\begin{equation}
\label{eq:variance-reduction}
\frac{
    \mathrm{Var}\!\bigl[\hat{g}_i^{\;\mathrm{split}}\bigr]
}{
    \mathrm{Var}\!\bigl[\hat{g}_i^{\;\mathrm{standard}}\bigr]
}
\;\approx\;
\frac{
    \bigl\lVert \nabla_{\theta_{\mathrm{trunk}}}\mathcal{L} \bigr\rVert^2
}{
    \bigl\lVert \nabla_{\theta}\mathcal{L} \bigr\rVert^2
}
\;\approx\;
\frac{P_{\mathrm{trunk}}}{P_{\mathrm{total}}}
\end{equation}
where the first relation is leading-order (it drops the $O(g_i^2)$ diagonal
terms; the exact per-coordinate identities are derived in
Appendix~\ref{app:variance-proof}) and the final approximation holds when
gradient magnitudes are approximately uniform across parameters. The trunk and
head parameters
partition the network, $P_{\mathrm{total}} = P_{\mathrm{trunk}} +
P_{\mathrm{head}}$, so the ratio
$P_{\mathrm{trunk}}/P_{\mathrm{total}} =
P_{\mathrm{trunk}}/(P_{\mathrm{trunk}} + P_{\mathrm{head}})$ is small precisely
when the head holds most of the weights.


Equation~\eqref{eq:variance-reduction} exposes the single lever that governs
Split-FG: the estimator is effective precisely when the forward-mode trunk holds
few parameters relative to the exactly differentiated head
($P_{\mathrm{trunk}} \ll P_{\mathrm{head}}$). Language models satisfy this for
free through their large vocabulary readouts. The same lever separates
Split-FG from the closest prior variance-reduction program:
\citet{ren2023scaling} also shrink the perturbed dimension, but with
layer-local losses at the price of a surrogate objective; Split-FG keeps the
global loss and removes the head's share exactly. The adversarial case
is a light head over a heavy trunk, which is most notably a plain convolutional
network making the variance ratio approaches one and Split-FG degrades to
standard forward gradient; the resolution is to restructure the network around
a small forward-mode feature extractor and a large exact head, the inversion we
adopt for image classification (Section~\ref{sec:image-experiment}). The same
light trunk also minimizes forward-mode activation memory
(Appendix~\ref{app:image-protocol}), so the heavy-head/light-trunk principle
improves accuracy and memory together.

\subsection{Parameter Update Rule}

Split-FG uses exact gradients for the head and forward-gradient estimates for
the trunk, then updates both parameter sets with Adam.

Let's first recall the gradient of head parameters. For a linear head $\mathrm{logits} = h\,W_{\mathrm{head}}^\top + b$ with
cross-entropy loss, the gradients are available in closed form:
\begin{equation}
\label{eq:head-grads}
\frac{\partial \mathcal{L}}{\partial b}
= \frac{1}{N}\sum_{n}\bigl(p_n - \mathbf{1}_{y_n}\bigr),
\qquad
\frac{\partial \mathcal{L}}{\partial W_{\mathrm{head}}}
= \tfrac{1}{N}\bigl(p - \mathbf{1}_{y}\bigr)^{\!\top} h,
\qquad
\frac{\partial \mathcal{L}}{\partial h}
= \tfrac{1}{N}\bigl(p - \mathbf{1}_{y}\bigr) W_{\mathrm{head}},
\end{equation}
Here $p = \mathrm{softmax}(\mathrm{logits})$, $\mathbf{1}_{y}$ is the one-hot
target matrix, and $N=BT$ is the number of token positions. These are the
usual one-layer head derivatives, computed directly rather than with an
autodiff tape. Split-FG eliminates reverse-mode differentiation through the
trunk, not through the head.

As to the trunk parameters, where the forward gradient comes into play, we compute a trunk JVP and
its directional derivative for $K$ random tangent vectors
$v_k \sim \mathcal{N}(0, I_{P_{\mathrm{trunk}}})$, :
\begin{equation}
\label{eq:trunk-grad-est}
\Delta h_k = \frac{\partial h}{\partial \theta_{\mathrm{trunk}}}\, v_k,
\qquad
d_k = \Bigl\langle
        \frac{\partial \mathcal{L}}{\partial h},\, \Delta h_k
      \Bigr\rangle,
\qquad
\hat{g}_{\mathrm{trunk}}
    = \frac{1}{K} \sum_{k=1}^{K} d_k \, v_k ,
\end{equation}
Each tangent requires one JVP and one inner product. The estimator is unbiased
because $\mathbb{E}[v v^\top]=I$.

Next, we use bias-corrected Adam \citep{kingma2017adammethodstochasticoptimization}, with the exact head
gradient and the trunk estimate in Eq.~\eqref{eq:trunk-grad-est}. Its
per-coordinate normalization handles the scale mismatch between head gradients
of magnitude $O(1)$ and trunk estimates with variance
$O(P_{\mathrm{trunk}}/K)$. However, it does not account for each coordinate's
signal-to-noise ratio, so noisy trunk coordinates can still receive overly
large updates. For large trunks, we therefore use a separate trunk step scale
(Section~\ref{sec:main-experiment}). Adam, AdamW, and Muon work in our
ablations, whereas SGD forward gradients remain at the random-prediction floor
on WikiText-103; even tuned pure FG with Adam is $7.5\times$ worse than
Split-FG (Table~\ref{tab:main-lm}).

\section{Convergence analysis for a predictable preconditioner}
\label{sec:theory}

In this section we will provide a theoretical support for the convergence of Split-FG under Adam. Throughout this section $F(\theta)$ denotes the full-batch training objective,
i.e.\ the loss $\mathcal{L}$ of Section~\ref{preliminary} evaluated on the full
training set. The Split-FG estimator of Section~\ref{split-fg} is conditionally
unbiased for $\nabla F(\theta)$ when the loss is evaluated full-batch;
minibatch sampling adds only the standard data-noise term, handled at the end
of Appendix~\ref{app:convergence-proofs}. For simplicity, we analyse the idealised update $\theta_{t+1}=\theta_t-\eta A_t\hat g_t$, where
$\hat g_t$ is the Split-FG estimator (exact head, $K$-tangent trunk) and $A_t$
is a diagonal preconditioner, under the following assumptions.

\begin{assumption}[Regularity]\label{ass:smooth}
$F$ is differentiable and $L$-smooth, and bounded below:
$F(\theta)\ge F_\star>-\infty$ for all $\theta$.
\end{assumption}

\begin{assumption}[Predictable, bounded preconditioner]\label{ass:precond}
$A_t=\mathrm{diag}(a_{t,1},\dots,a_{t,P_{\mathrm{total}}})$ is
\emph{predictable}---measurable with respect to the history $\mathcal{F}_t$
fixed before the step-$t$ tangents are drawn---and satisfies
$0<a_{\min}\le a_{t,i}\le a_{\max}<\infty$ for all $t,i$.
\end{assumption}

\begin{assumption}[Unbiasedness and bounded adaptive noise]\label{ass:noise}
The estimator is conditionally unbiased,
$\mathbb{E}_t[\hat g_t]=\nabla F(\theta_t)$, and its adaptive noise ratio
$R_t(A_t)=\mathbb{E}_t[\lVert A_t\hat g_t\rVert^2]\big/
\langle\nabla F(\theta_t),A_t\nabla F(\theta_t)\rangle$
(set to $0$ when $\nabla F(\theta_t)=0$) obeys $R_t(A_t)\le R_{\max}$ almost
surely.
\end{assumption}

\begin{theorem}[Predictable-preconditioner Split-FG]\label{thm:adam}
Under Assumptions~\ref{ass:smooth}--\ref{ass:noise}, if
$0<\eta\le \frac{1}{L R_{\max}}$ then for every horizon $T\ge1$
\begin{equation*}
\frac{1}{T}\sum_{t=0}^{T-1}
\mathbb{E}\big[\langle\nabla F(\theta_t),A_t\nabla F(\theta_t)\rangle\big]
\;\le\;\frac{2\,(F(\theta_0)-F_\star)}{\eta\,T},
\end{equation*}
and consequently
$\tfrac{1}{T}\sum_{t=0}^{T-1}\mathbb{E}\lVert\nabla F(\theta_t)\rVert^2
\le 2(F(\theta_0)-F_\star)/(\eta\,a_{\min}T)$. The Split-FG estimator satisfies
$R_{\max}\le(a_{\max}^2/a_{\min})\big(1+(P_{\mathrm{trunk}}+1)/K\big)$, so the
step-size condition is always feasible; the guarantee is the standard $O(1/T)$
rate for nonconvex first-order stationarity.
\end{theorem}

Theorem~\ref{thm:adam} certifies descent in the geometry chosen by the
preconditioner: the stationarity measure is the preconditioned gradient energy
$\langle\nabla F,A_t\nabla F\rangle$, and the head coordinates contribute no
variance term because they are updated exactly. It guarantees convergence, not
the \emph{quality} of the point reached within a finite budget---the
second-moment denominator equalises update magnitudes but does not distinguish
signal from noise, and Section~\ref{sec:main-experiment} shows that at large
$P_{\mathrm{trunk}}$ a further trunk-specific step scale $\rho$ is needed. That
variant is covered unchanged, since scaling the trunk entries of $A_t$ by a
constant $\rho$ preserves Assumption~\ref{ass:precond}. The proof, the exact
weighted second moment underlying the $R_{\max}$ bound, and an unpreconditioned
SGD counterpart are in Appendix~\ref{app:convergence-proofs}.

\section{Experiments}
\label{sec:experiments}

This section is designed to test Split-FG as a general training principle,
not only as a language-model trick. We organize the experiments around four
questions. First, does the estimator variance follow the scaling predicted
by Eq.~\eqref{eq:variance-reduction}? Second, does Split-FG work with
different loss functions, including cross-entropy classification and MSE
regression? Third, does the method transfer across modern tabular, vision,
and language architectures? Fourth, which design choices matter most in
GPT-style transformers?

\paragraph{Protocol and metrics.}
Tabular and GPT experiments pair Split-FG with a same-architecture
backpropagation reference under the same data-exposure budget. The headline
image table instead compares separately structured systems and, on CIFAR-100,
method-specific budgets; its architecture-matched flat-head ablation is reported
separately. Each large experiment reports peak GPU memory together with the
reference difference
$\mathrm{MemSave}=1-M_{\mathrm{peak}}(\text{Split-FG})/M_{\mathrm{peak}}(\text{Backprop})$.
The full training and memory-measurement protocol is given in
Appendix~\ref{app:general-protocol}.

\paragraph{No-global-backward baselines.}
Beyond pure forward gradient and antithetic ES, we compare---wherever they
apply against Forward-Forward
\citep{hinton2022forwardforwardalgorithmpreliminaryinvestigations} (layer-local
goodness training; classification only), predictive coding
\citep{Whittington2017AnAO,millidge2022predictive} (local error settling with
Hebbian updates), and a training-free modern-Hopfield associative memory
\citep{ramsauer2021hopfield} (softmax retrieval over a frozen trunk; kNN-LM
style \citep{khandelwal2020knnlm} for language modelling). All three use zero
JVPs and zero trunk backward passes; their mechanisms and per-domain
configurations are detailed in Appendix~\ref{app:general-protocol} and the
per-domain protocol appendices.

\subsection{Toy Illustration: Variance at the Split}
\label{sec:toy-experiment}

We first construct a toy model in which the total parameter vector can be
partitioned into a large linear head and a small nonlinear trunk. The model
has the form
\begin{equation}
    h = f(x;\theta_{\mathrm{trunk}}), \qquad
    \hat{y} = W_{\mathrm{head}}h + b ,
\end{equation}
with squared loss or cross-entropy loss depending on the output type. This
setting allows us to measure the true gradient exactly and compare the
empirical variance of three estimators: standard forward gradient over all
parameters, Split-FG with an exact head gradient, and backpropagation. We
vary the head dimension while holding the trunk fixed, so that
$P_{\mathrm{head}}/P_{\mathrm{trunk}}$ can be swept directly.

We instantiate this diagnostic with a two-layer ReLU MLP trunk
($P_{\mathrm{trunk}}=1600$) and a linear cross-entropy head whose class count
$C\in\{50,500,5000\}$ grows the head from $1650$ to $165000$ parameters at
fixed trunk; for each $C$ we draw $2000$ single-direction ($K{=}1$) estimates
against the exact minibatch gradient and report variance over the trunk
coordinates (the analytical head contributes none). Full protocol in
Appendix~\ref{app:variance-proof}.

For Gaussian tangents, the leading-order variance-ratio prediction on a fixed
minibatch is
\begin{equation}
\label{eq:toy-energy-ratio}
    \frac{
        \mathrm{Var}[\hat{g}^{\mathrm{split}}_{\mathrm{trunk}}]
    }{
        \mathrm{Var}[\hat{g}^{\mathrm{pure}}_{\mathrm{trunk}}]
    }
    =
    \frac{
        \|\nabla_{\theta_{\mathrm{trunk}}}\mathcal{L}\|_2^2
    }{
        \|\nabla_{\theta}\mathcal{L}\|_2^2
    } .
\end{equation}
The parameter-count ratio $P_{\mathrm{trunk}}/P_{\mathrm{total}}$ is thus a
uniform-gradient-energy approximation, not an identity. This distinction
matters in the toy classifier: as the number of classes grows, the count
ratio becomes very small, but the observed ratio follows the
gradient-energy prediction.

The toy result (Figure~\ref{fig:toy-variance} and
Table~\ref{tab:toy-variance}, Appendix~\ref{app:variance-proof})
supports the core variance mechanism while also clarifying
what should be compared in finite models. Split-FG consistently reduces the
trunk-coordinate variance relative to pure forward gradient, and the
measured ratios are within $6.8\%$ of Eq.~\eqref{eq:toy-energy-ratio} across
the sweep. Backpropagation is included only as the zero-variance reference:
it supplies the exact gradient used to evaluate estimator variance, but it is
not a stochastic estimator.

\subsection{Tabular Learning with TabM-Style Models}
\label{sec:tabular-experiment}

The tabular experiments test both architectural generality and loss-function
generality. We use a TabM-style parameter-efficient ensemble model, with
feature preprocessing, a shared MLP trunk, and multiple prediction heads or
submodel outputs. For classification datasets, the final layer uses
cross-entropy and the analytical head gradient from
Eq.~\eqref{eq:head-grads}. For regression datasets, the final layer uses MSE,
whose closed-form head gradient mirrors Eq.~\eqref{eq:head-grads}
(Appendix~\ref{app:tabular-arch}).

The dataset subset follows the spirit of the TabM \citep{gorishniy2025tabm} benchmark while keeping
the experiment time reasonable; the datasets are drawn mainly from OpenML. Full dataset and protocol details are in Appendix~\ref{datasets}, and the network structure in Appendix~\ref{app:tabular-arch}.

\begin{table}[h]
\caption{Tabular benchmarks: held-out-fold accuracy (\%, classification) and
RMSE (regression), mean $\pm$ std over $5$-fold cross-validation; ``--'' marks
methods that do not apply. Protocol and model structure:
Appendices~\ref{app:tabular-protocol} and~\ref{app:tabular-arch}.}
\label{tab:tabular-main}
\begin{center}
\small
\resizebox{\textwidth}{!}{%
\begin{tabular}{llcccccccc}
\hline
Dataset & Task & Loss & Backprop & Pure FG & ES & FF & PC & Hop. & Split-FG \\
\hline
Adult                 & bin. cls. & CE  & $84.2_{\pm0.5}$ & $81.7_{\pm0.1}$ & $81.8_{\pm0.1}$ & $81.6_{\pm0.5}$ & $81.5_{\pm0.4}$ & $79.1_{\pm0.4}$ & $\textbf{82.5}_{\pm0.4}$ \\
Higgs Small           & bin. cls. & CE  & $68.8_{\pm0.7}$ & $59.0_{\pm0.6}$ & $59.0_{\pm0.6}$ & $61.7_{\pm0.4}$ & $61.5_{\pm0.4}$ & $55.4_{\pm0.3}$ & $\textbf{61.9}_{\pm0.3}$ \\
Otto   & multicls. & CE  & $78.3_{\pm0.4}$ & $62.8_{\pm0.6}$ & $62.8_{\pm0.6}$ & $30.8_{\pm3.5}$ & $68.8_{\pm0.4}$ & $62.3_{\pm0.7}$ & $\textbf{71.7}_{\pm0.4}$ \\
Covertype             & multicls. & CE  & $71.7_{\pm0.5}$ & $61.2_{\pm0.2}$ & $61.2_{\pm0.2}$ & $35.0_{\pm2.2}$ & $65.5_{\pm0.6}$ & $62.1_{\pm1.1}$ & $\textbf{66.8}_{\pm0.6}$ \\
California     & reg.      & MSE & $0.491_{\pm0.013}$ & $0.595_{\pm0.012}$ & $0.595_{\pm0.012}$ & -- & $0.638_{\pm0.015}$ & $0.755_{\pm0.014}$ & $\textbf{0.564}_{\pm0.016}$ \\
House 16H             & reg.      & MSE & $0.652_{\pm0.024}$ & $0.829_{\pm0.027}$ & $0.829_{\pm0.027}$ & -- & $0.838_{\pm0.033}$ & $0.859_{\pm0.040}$ & $\textbf{0.773}_{\pm0.030}$ \\
Diamond               & reg.      & MSE & $0.249_{\pm0.004}$ & $0.334_{\pm0.006}$ & $0.334_{\pm0.006}$ & -- & $0.452_{\pm0.001}$ & $0.490_{\pm0.003}$ & $\textbf{0.311}_{\pm0.004}$ \\
Black Friday          & reg.      & MSE & $0.809_{\pm0.006}$ & $0.906_{\pm0.003}$ & $0.906_{\pm0.003}$ & -- & $0.901_{\pm0.004}$ & $0.938_{\pm0.005}$ & $\textbf{0.890}_{\pm0.006}$ \\
\hline
\end{tabular}}
\end{center}
\end{table}

Across all eight default datasets, Split-FG consistently improves over the
evaluated pure forward-gradient and ES controls under $5$-fold
cross-validation (paired $t$-tests)---e.g.\ on Otto it lifts
accuracy from $62.8\%$ to $71.7\%$ against a $78.3\%$ backprop reference. On
Higgs its lead over Forward-Forward is within the cross-validation spread.
Split-FG also lowers peak training memory on every dataset measured
($4.3$--$24.1\%$; Table~\ref{tab:tabular-memory} in the appendix).

\subsection{Image Classification}
\label{sec:image-experiment}

Convolutional networks are the adversarial case for Split-FG
(Section~\ref{split-fg}): a ResNet holds almost all of its parameters in the
convolutional trunk and only a thin linear head, so a naive split must
forward-estimate the whole trunk and degrades to standard forward gradient. We
therefore apply the heavy-head/light-trunk restructuring: for CIFAR-10 we keep
only the first residual block of a ResNet-20 (GroupNorm) as the forward-mode
trunk ($5{,}136$ parameters, under $2\%$ of the backbone) and attach a large
exact head that flattens its feature map and maps it to the logits through a
two-layer \emph{linear} factorisation,
\begin{equation}
    h = \mathrm{ResBlock}_1(x;\theta_{\mathrm{trunk}}) \in \mathbb{R}^{16384},
    \qquad
    \mathrm{logits} = W_2 (W_1 h) + b ,
\end{equation}
trained by an exact closed-form gradient (pure matrix products, no reverse-mode
backward pass through the trunk). Although $W_2 W_1$
collapses to a single linear map, training the factored pair yields a deep-linear
over-parameterisation effect worth $+2.5$ points over a single exact linear head
($p<0.01$; Appendix~\ref{app:image-ablation}). The inner width is $H{=}64$, so
$W_1\in\mathbb{R}^{H\times16384}$ and $W_2\in\mathbb{R}^{C\times H}$ for $C$
classes; we use $K{=}4$ tangents ($H{=}128$ for CIFAR-100, which uses a stronger
ResNet-18 backbone). The baselines use the dataset's standard
backbone: Pure FG and ES estimate the full network, Forward-Forward (FF) and
predictive coding (PC) run on flattened images, and Hopfield (Hop.) is an
associative-memory readout over a frozen trunk. Dataset, protocol, and
architecture details are given in
Appendices~\ref{app:image-datasets}--\ref{app:image-arch}.

\begin{table}[t]
\caption{Image development evaluation: top-1 accuracy (\%).  Protocol and model structure:
Appendices~\ref{app:image-protocol} and~\ref{app:image-arch}.}
\label{tab:image-main}
\begin{center}
\small
\resizebox{\textwidth}{!}{%
\begin{tabular}{lccccccc}
\hline
Dataset & Backprop & Pure FG & ES & FF & PC & Hop. & Split-FG \\
\hline
CIFAR-10     & $88.2_{\pm0.6}$ & $29.5_{\pm1.9}$ & $29.3_{\pm1.4}$ & $48.8_{\pm0.4}$ & $40.7_{\pm1.1}$ & $10.0_{\pm0.0}$ & $60.5_{\pm1.2}$ \\
CIFAR-100    & $65.6_{\pm1.0}$ & $1.2_{\pm0.2}$ & $1.2_{\pm0.1}$ & $6.9_{\pm0.5}$ & $1.0_{\pm0.0}$ & $1.4_{\pm0.3}$ & $35.2_{\pm0.5}$ \\
\hline
\end{tabular}}
\end{center}
\end{table}

In the reported five-seed CIFAR-10 development results, the restructured Split-FG system
reaches $60.5\%$ after $60$ epochs (Table~\ref{tab:image-main}), compared with
$29.5\%$ for full-ResNet pure FG and $48.8\%$ for the separately structured
Forward-Forward system; because these methods use different architectures,
this does not isolate an estimator effect. Split-FG is $27.7$ points below the
full ResNet-20 backpropagation reference ($88.2\%$). Published
no-global-backward pipelines with heavier machinery reach far higher. For example, DeepZero
\citep{chen2024deepzero} attains ${\sim}86\%$ on CIFAR-10 with sparse
zeroth-order training of a full ResNet-20 at substantially larger compute.
Table~\ref{tab:image-main} should therefore be read as preliminary system
evidence, not a state-of-the-art or matched-method comparison.
The exact head is what makes the split work: forward-estimating it instead
collapses accuracy from $56.9\%$ to $\approx\!22\%$ in the matched ablation of
Appendix~\ref{app:image-ablation}. The light trunk also cuts peak memory to
$1282$\,MiB, below backpropagation's $1520$\,MiB and reversing the
$2668$\,MiB blow-up of the naive full-trunk split
(Table~\ref{tab:image-memory}). On CIFAR-100, the $60$-epoch Split-FG system
reaches $35.2_{\pm0.5}\%$ and the $60$-epoch ResNet-18 backpropagation reference
reaches $65.6_{\pm1.0}\%$. The remaining methods were stopped at the same
budgets, so their numbers  support a matched-budget
ranking.

\subsection{GPT-Style Language Modelling}
\label{sec:main-experiment}

The main stress test is GPT-2-style causal language modelling from random
initialisation on WikiText-103, a standard real-text benchmark. The model is a causal transformer (token and positional embeddings, GPT
blocks, final normalisation, linear LM head); the default split is before the
LM head:
\begin{equation}
    h = \mathrm{GPT}_{\mathrm{trunk}}(x;\theta_{\mathrm{trunk}}),
    \qquad
    \mathrm{logits}=hW_{\mathrm{out}}^\top+b .
\end{equation}
We report single-seed validation NLL and perplexity, mean step time, and peak GPU memory;
the tangent-count ablation in Appendix~\ref{app:gpt-k-ablation} additionally
reports JVPs per step. The same-architecture Adam backpropagation and Split-FG runs use
the same number of training tokens and optimizer updates. The concrete
\texttt{small8} architecture ($4$ layers, $d_{\mathrm{model}}=256$, $16.1$M
parameters) is listed in Table~\ref{tab:gpt-arch}, and the full protocol in
Appendix~\ref{app:gpt-protocol}.

Table~\ref{tab:main-lm} compares three Split-FG settings: a naive run that
uses Adam's full step for the trunk (perplexity $1733.4$), a frozen-trunk
control that trains only the exact head ($667.6$), and a run with the trunk
step scaled by $\rho=0.03$ ($386.7$). Unless noted otherwise, FG and ES use
one tangent. The $K{=}4$ pure-FG baseline is learning-rate tuned using the
screening protocol in Appendix~\ref{app:gpt-protocol}. The backpropagation
reference uses Adam without weight decay; a separately tuned AdamW run reaches
$75.5$ perplexity. Hopfield time denotes a full retrieval pass.

The naive run performs worse than freezing the trunk. We believe this
is because the unbiased trunk estimate still has very low per-coordinate
signal-to-noise ratio (${\sim}1/P_{\mathrm{trunk}}$; Eq.~\eqref{eq:snr}). Adam
therefore gives full-sized updates to many weakly determined trunk coordinates,
which can cause the trunk to drift before its signal accumulates. Reducing the
trunk step by $\rho\!\ll\!1$ limits this noise-driven drift while preserving
the slower, coherent update. At matched $K{=}4$, this simple change improves
perplexity by $4.5\times$ with the same peak memory and essentially unchanged
step time. Figure~\ref{fig:frozen-control} shows the training curves; further
frozen-trunk and trunk-scale controls are in Appendix~\ref{sec:frozen-control}.

\begin{table}[t]
\caption{ WikiText-103 validation results for the
\texttt{small8} GPT model. All neural methods use one epoch of matched token
exposure. Hyperparameters were selected using validation results; no test-set
claim is made. Lower perplexity is better.}
\label{tab:main-lm}
\begin{center}
\small
\resizebox{\linewidth}{!}{%
\begin{tabular}{lccccc}
\hline
Method & Train exposure & Val. NLL & PPL & Time & Peak mem.  \\
\hline
Same-architecture backprop + Adam & \(119.1\)M / \(1\) ep. & \(\mathbf{5.0080}\) & \(\mathbf{149.61}\) & \(41.7\) ms & \(1237.7\) MiB  \\
Forward gradient + Adam, \(K{=}1\) & same                   & \(10.5999\) & \(40128.90\)      & \(63.6\) ms & \(1038.4\) MiB  \\
Forward gradient + Adam, \(K{=}4\), tuned lr & same         & \(7.9673\) & \(2884.97\)        & \(137.4\) ms & \(1694.5\) MiB  \\
Forward gradient + SGD, \(K{=}1\)  & same                   & \(10.8274\) & \(50385.05\)      & \(54.6\) ms & \(913.7\) MiB  \\
Antithetic ES + Adam, \(K{=}1\)    & same                   & \(9.4012\) & \(12103.11\)       & \(44.1\) ms & \(764.7\) MiB   \\
Hopfield assoc.\ memory (kNN-LM)  & \(4096\) windows       & \(7.4466\) & \(1714.02\)        & \(48.5\) s  & \(5368.1\) MiB  \\
Unigram reference (count-based)   & train counts           & \(7.4140\) & \(1659.03\)        & ---         & ---             \\
Interpolated bigram reference     & train counts           & \(5.2889\) & \(198.12\)         & ---         & ---             \\
Split-FG + Adam, \(K{=}4\), naive trunk step (\(\times1\)) & \(119.1\)M / \(1\) ep. & \(7.4578\) & \(1733.36\) & \(137.3\) ms & \(810.6\) MiB \\
Frozen trunk $+$ exact head (\(K{=}0\)) & \(119.1\)M / \(1\) ep. & \(6.5037\) & \(667.64\) & \(24.4\) ms & \(810.6\) MiB \\
Split-FG + Adam, \(K{=}4\), trunk step \(\times0.03\) & \(119.1\)M / \(1\) ep. & \(5.9577\) & \(386.70\) & \(133.8\) ms & \(810.6\) MiB \\
\hline
\end{tabular}
}
\end{center}
\end{table}

\begin{figure}[t]
\begin{center}
\includegraphics[width=0.46\linewidth]{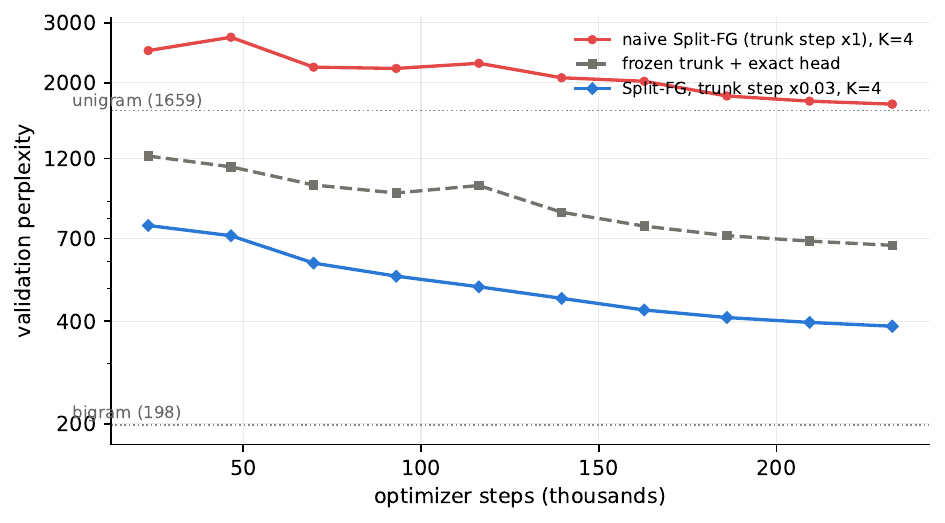}
\end{center}
\caption{Frozen-trunk control, naive Split-FG ($K{=}4$), and $\rho$-scaled
Split-FG on WikiText-103, same one-epoch cosine schedule (single seed).
Naive trunk training tracks \emph{above} the frozen control throughout; the
$\rho$-scaled trunk drops below it from the first checkpoint and passes the
frozen run's final perplexity in about a quarter of the budget.}
\label{fig:frozen-control}
\end{figure}

\paragraph{Results.} In this single-seed validation comparison, the
same-architecture backpropagation reference (Adam, no weight decay, as for
Split-FG) reaches perplexity $149.6$; Split-FG with a reduced trunk step reaches
$386.7$ (Appendix~\ref{sec:frozen-control}). It performs better than the
evaluated trainable no-global-backward baselines in this table. The comparison that most directly isolates the split
is pure forward gradient at the same $K{=}4$ and with the same learning-rate
screening procedure (Appendix~\ref{app:gpt-protocol}): it reaches $2{,}885$, so
removing the split, which estimates the
head instead of solving it exactly, costs $7.5\times$ at equal $K$ (and
$104\times$ against the untuned $K{=}1$ default). The training-free
Hopfield/kNN-LM baseline holds $6.6\times$ Split-FG's memory.
The count-based unigram/bigram references are fitted on the training tokens
and evaluated identically; the interpolated bigram ($198$)
still leads all evaluated trainable no-global-backward methods
(Appendix~\ref{sec:frozen-control}). Split-FG trains in
$810.6$\,MiB against $1237.7$\,MiB for backpropagation
($\mathrm{MemSave}=34.5\%$, Eq.~\eqref{eq:memory-saving}). The costs are a
$3.2\times$ step time and a $2.6\times$ perplexity gap to backpropagation,
which come from estimating the trunk with $K$ scalar probes per step.

\subsection{GPT Ablation Studies}
\label{sec:ablation}

We isolate each design choice by changing one component at a time from the
main configuration (full tables in
Appendices~\ref{app:gpt-core-ablation}--\ref{app:gpt-k-ablation}).
Removing the trunk step scale costs $4.5\times$ in perplexity, freezing the
trunk costs $1.7\times$, and dropping the split entirely costs $7.5\times$
against a pure-FG baseline matched in tangent count and tuning ($104\times$
against the untuned $K{=}1$ default). These single-seed controls are consistent
with the split removing a dominant noise source and the reduced step making the
remaining trunk estimate usable. In the reported $100\times$ grid of $\rho$,
every value beats the frozen control at both the
screening and the full budget, with a broad minimum around
$\rho\in[0.03,0.1]$ (Table~\ref{tab:rho-ablation};
Appendix~\ref{app:gpt-rho}). Increasing $K$ from $1$
to $8$ improves perplexity monotonically at constant peak memory (the $K$
tangents run sequentially, each freeing its tape before the next) and
sub-linear wall-clock cost ($52\to187$\,ms per step), making $K$ a pure
time--variance dial.

\begin{table}[t]
\caption{Trunk-step-scale sensitivity on WikiText-103 (\texttt{small8},
Split-FG + Adam, $K{=}4$, single seed) at two fully annealed budgets:
$10\%$-epoch screening (matched frozen control $816.0$) and full epoch
(frozen $667.6$). Every $\rho$ beats the frozen control at both budgets; the
optimum $\rho=0.1$ sits one grid step above the paper's default $\rho=0.03$,
whose full-budget entry is the main-text run (Table~\ref{tab:main-lm}).}
\label{tab:rho-ablation}
\begin{center}
\small
\begin{tabular}{lcc}
\hline
Trunk step scale $\rho$ & $10\%$-budget PPL & Full-budget PPL \\
\hline
$0.003$ & $716.1$ & $622.7$ \\
$0.01$  & $681.8$ & $514.9$ \\
$0.03$  & $623.3$ & $386.7$ \\
$0.1$   & $541.9$ & $365.0$ \\
$0.3$   & $677.9$ & $633.1$ \\
\hline
Frozen control ($K{=}0$) & $816.0$ & $667.6$ \\
\hline
\end{tabular}
\end{center}
\end{table}
Among optimizers (Appendix~\ref{app:gpt-optim}), the three adaptive methods
cluster tightly (Muon $382$, Adam $387$, AdamW $399$) while SGD fails
entirely ($44{,}918$): only a per-coordinate denominator can reconcile the
$O(1)$ head gradient with the $O(P_{\mathrm{trunk}}/K)$ trunk estimate under
one learning rate.


\section{Discussion}
\label{discussion}

\paragraph{Limitations.}
Split-FG still trails backpropagation: on WikiText-103 it reaches perplexity
$387$, versus $150$ with matched Adam and $75.5$ with tuned AdamW, with gaps
of $27.7$ and $30.4$ points on CIFAR-10 and CIFAR-100. Its variance also
favors architectures with a large exact head; our flat-head image models make
this trade-off by using a shallow feature extractor. Memory savings come at a
substantial time cost ($1.5$--$4.5\times$ slower steps for $K{=}2$--$8$), and
activation checkpointing \citep{chen2016trainingdeepnetssublinear} or reversible architectures
\citep{gomez2017reversible} remain stronger choices when memory is the only
goal.  Split-FG's distinct property is eliminating reverse-mode passes through
the trunk.This approach is especially valuable where a backward pass is unavailable, such as on the optical and analog accelerators motivating this work, as it obtains trunk gradients solely through forward perturb-and-measure techniques Our experiments are limited to a $16$M-parameter language model and
vision backbones up to $11$M parameters on one 8\,GB GPU; because variance
scales as $P_{\mathrm{trunk}}/K$, larger trunks remain the main open question.
The GPT results are single-seed, use gradient clipping and a readout-only
surrogate for tied embeddings (Appendices~\ref{app:general-protocol}
and~\ref{app:gpt-arch}). Comparisons across $K$ are additionally affected by separate annealing
schedules.

\paragraph{A negative result: sizing the trunk to the probe budget.}
We first tested whether reducing the forward-trained trunk to match the probe
budget would help. This ``light trunk'' trains only positional embeddings,
LayerNorm affines, and biases ($46.6$k parameters, roughly $20$ probes per
parameter), while freezing the trunk weight matrices. It only ties the frozen
control ($671.6$ vs.\ $667.6$ perplexity), whereas the $\rho$-scaled full trunk
reaches $386.7$. The limiting factor is therefore step size, not probe count:
with a reduced trunk step, even a severely under-determined trunk can
learn. Here $\rho$ is simply a per-group learning-rate multiplier
\citep{you2020largebatchoptimizationdeep,howard-ruder-2018-universal,yang2021tuning}; the key
point is that the noisy, estimated trunk gradient needs a much smaller step
than the exact head gradient.

\paragraph{Future directions.}
Two extensions stand out: conditioned tangents, which scale each tangent
coordinate by a running gradient-magnitude estimate, reweighted to stay
unbiased (preliminary runs: better sample efficiency, same plateau under full
annealing) and broader head classes, e.g.\ a head-only backward pass
for nonlinear heads or finite-difference head gradients for black-box heads.



\bibliography{iclr2026_conference}
\bibliographystyle{iclr2026_conference}

\appendix

\section{Related Work}
\label{sec:related}

\paragraph{Forward gradients and zeroth-order training.}
Forward-mode directional derivatives as unbiased gradient estimates were
proposed by \cite{baydin2022gradientsbackpropagation} and studied for
representation learning by \citet{silver2022learning}; evolution strategies
\citep{salimans2017evolutionstrategiesscalablealternative,JMLR:v15:wierstra14a}
and zeroth-order fine-tuning \citep{malladi2023finetuning} estimate the same
object from function evaluations. All inherit variance that grows with the
number of perturbed parameters. \citet{ren2023scaling} reduce that dimension
with layer-local losses at the price of a surrogate objective; Split-FG instead
keeps the \emph{global} loss and removes the head's share of the dimension
exactly, which is where most parameters live in our settings. Our frozen-trunk
control asks a question this literature rarely does---whether the estimated
component improves on \emph{not training it at all}---and our answer (no,
unless the update is rescaled) suggests such controls should be standard.

\paragraph{Local and biologically motivated learning.}
Forward-Forward \citep{hinton2022forwardforwardalgorithmpreliminaryinvestigations},
predictive coding \citep{Whittington2017AnAO,millidge2022predictive}, equilibrium
propagation \citep{EP}, and feedback alignment \citep{Nkland2016DirectFA}
avoid the backward pass with local objectives or fixed feedback, introducing
bias; associative-memory readouts \citep{ramsauer2021hopfield,khandelwal2020knnlm}
avoid training altogether. We compare against these families directly
(Section~\ref{sec:experiments}).

\paragraph{Per-group step sizes.}
Mechanically, our trunk step scale $\rho$ is a per-group learning-rate
multiplier---an axis long used by layer-wise adaptive methods
\citep{you2017lars,you2020largebatchoptimizationdeep}, discriminative fine-tuning
\citep{howard2018universal}, and width-aware parameterisations
\citep{yang2021tuning}. We claim no novelty for the mechanism. The
contribution is the \emph{diagnosis}: with a mixed exact/estimated gradient,
Adam-style normalisation equalises update magnitudes but not signal-to-noise,
so the estimated group must be slowed by orders of magnitude ($\rho\approx
0.03$ at $P_{\mathrm{trunk}}{=}3.2$M)---without which trunk training is worse
than leaving the trunk frozen, a failure mode invisible unless the frozen
control is run.

\paragraph{Memory-efficient exact-gradient training.}
Activation checkpointing \citep{chen2016trainingdeepnetssublinear} and reversible architectures
\citep{gomez2017reversible} reduce activation memory while keeping exact
gradients, and on pure memory grounds they are strong alternatives to
forward-mode training. Split-FG's claim is different: it uses \emph{zero}
reverse-mode passes through the trunk, the regime relevant to hardware that
cannot run a backward pass and to models of biological learning; its memory
saving (Section~\ref{sec:experiments}) is a by-product, not the primary
argument.

\section{Variance Reduction Derivation}
\label{app:variance-proof}

Let $\theta=(\theta_{\mathrm{trunk}},\theta_{\mathrm{head}})$ and write
$F(\theta)$ for the full-batch training objective. At a fixed iterate, denote
the exact gradients by
$g_{\mathrm{trunk}}=\nabla_{\theta_{\mathrm{trunk}}}F(\theta)$ and
$g_{\mathrm{head}}=\nabla_{\theta_{\mathrm{head}}}F(\theta)$. Split-FG draws
$K$ independent Gaussian tangent vectors
$v_k\sim\mathcal{N}(0,I_{P_{\mathrm{trunk}}})$ only in the trunk parameter
space and uses
\begin{equation}
\label{eq:split-estimator-analysis}
\hat g_{\mathrm{trunk}}
  = \frac{1}{K}\sum_{k=1}^K
    \langle g_{\mathrm{trunk}},v_k\rangle v_k,
\qquad
\hat g_{\mathrm{head}} = g_{\mathrm{head}} .
\end{equation}
Unbiasedness follows from $\mathbb{E}[v_kv_k^\top]=I$:
$\mathbb{E}[\langle g_{\mathrm{trunk}},v_k\rangle v_k]
=g_{\mathrm{trunk}}$.

For a single tangent and a trunk coordinate $i$, write
$\hat g_i=(\sum_j g_jv_j)v_i$. Since
$\mathbb{E}[v_jv_\ell v_i^2]=\delta_{j\ell}
+2\delta_{ij}\delta_{i\ell}$ for Gaussian $v$,
$\mathbb{E}[\hat g_i^2]=\lVert g_{\mathrm{trunk}}\rVert^2+2g_i^2$.
Subtracting the squared mean $g_i^2$ and averaging $K$ independent samples gives
\begin{equation}
\label{eq:split-coordinate-var}
\mathrm{Var}\!\left[(\hat g_{\mathrm{trunk}})_i\right]
  =
  \frac{1}{K}
  \left(
    \lVert g_{\mathrm{trunk}}\rVert^2
    +
    (g_{\mathrm{trunk}})_i^2
  \right).
\end{equation}
Thus Split-FG replaces the pure forward-gradient dominant term
$\lVert\nabla F(\theta)\rVert^2$ by $\lVert g_{\mathrm{trunk}}\rVert^2$ for
each trunk coordinate; the head contribution is exact and contributes no
estimator variance.

Summing over the $P_{\mathrm{trunk}}$ trunk coordinates gives
\begin{align}
\label{eq:split-total-var}
\mathbb{E}\!\left[\hat g\mid \theta\right]
  &= \nabla F(\theta),\\
\mathbb{E}\!\left[
  \lVert \hat g-\nabla F(\theta)\rVert^2
  \mid \theta
\right]
  &=
  \frac{P_{\mathrm{trunk}}+1}{K}
  \lVert g_{\mathrm{trunk}}\rVert^2,\\
\mathbb{E}\!\left[
  \lVert \hat g\rVert^2
  \mid \theta
\right]
  &=
  \lVert g_{\mathrm{head}}\rVert^2
  +
  \left(1+\frac{P_{\mathrm{trunk}}+1}{K}\right)
  \lVert g_{\mathrm{trunk}}\rVert^2 .
\end{align}
For pure forward gradient over all $P_{\mathrm{total}}$ parameters, the
corresponding total variance is
$((P_{\mathrm{total}}+1)/K)\lVert\nabla F(\theta)\rVert^2$. Under roughly
uniform gradient energy per coordinate, the dominant per-coordinate variance
ratio becomes $P_{\mathrm{trunk}}/P_{\mathrm{total}}$.

\begin{figure}[h]
\begin{center}
\includegraphics[width=\linewidth]{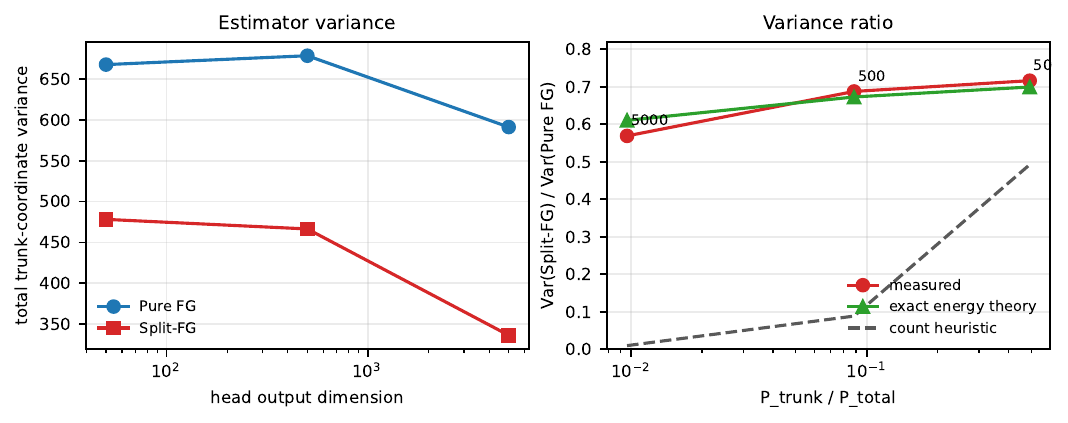}
\end{center}
\caption{Controlled toy variance experiment
(Section~\ref{sec:toy-experiment}). Left: total empirical variance
over trunk coordinates for pure forward gradient and Split-FG as the linear
head grows. Right: the measured variance ratio tracks the
gradient-energy prediction in Eq.~\eqref{eq:toy-energy-ratio}; the
parameter-count ratio is a useful heuristic only when gradient energy is
approximately uniform across parameters.}
\label{fig:toy-variance}
\end{figure}

\begin{table}[h]
\caption{Toy variance diagnostic with a fixed trunk and growing linear
classification head. Ratios use trunk-coordinate estimator variance; relative
error is with respect to the energy-ratio prediction of
Eq.~\eqref{eq:toy-energy-ratio}.}
\label{tab:toy-variance}
\begin{center}
\small
\begin{tabular}{rrrrrr}
\hline
Classes & $P_{\mathrm{total}}$ & $P_{\mathrm{trunk}}/P_{\mathrm{total}}$ &
Measured & Energy theory & Rel. error \\
\hline
50   & 3250   & 0.4923 & 0.7162 & 0.6993 & 2.4\% \\
500  & 18100  & 0.0884 & 0.6875 & 0.6727 & 2.2\% \\
5000 & 166600 & 0.0096 & 0.5689 & 0.6104 & 6.8\% \\
\hline
\end{tabular}
\end{center}
\end{table}

\section{Convergence Proofs}
\label{app:convergence-proofs}

\paragraph{Formal statement (predictable-preconditioner Split-FG).}
Let $F$ denote the full-batch training objective. Assume that $F$ is bounded
below by $F_\star$ and is $L$-smooth. Let $\hat g_t$ be the Split-FG estimator
at step $t$, with an exact head component and a $K$-sample forward-gradient
trunk component. Consider the diagonal-preconditioned update
\begin{equation}
\label{eq:adam-style-update}
\theta_{t+1}=\theta_t-\eta A_t \hat g_t,
\qquad
A_t=\mathrm{diag}(a_{t,1},\ldots,a_{t,P_{\mathrm{total}}}),
\end{equation}
where $A_t$ is predictable given the history before drawing the current
tangents and is uniformly bounded:
\begin{equation}
\label{eq:adam-preconditioner-bounds}
0<a_{\min}\leq a_{t,i}\leq a_{\max}<\infty .
\end{equation}
Define the adaptive noise ratio
\begin{equation}
\label{eq:adaptive-noise-ratio}
R_t(A_t)=
\frac{
  \mathbb{E}_t[\lVert A_t\hat g_t\rVert^2]
}{
  \langle \nabla F(\theta_t),A_t\nabla F(\theta_t)\rangle
}.
\end{equation}
When $\nabla F(\theta_t)=0$, both numerator and denominator vanish for the
full-batch estimator and we set $R_t(A_t)=0$ by convention.
If $R_t(A_t)\leq R_{\max}$ almost surely and
$0<\eta\leq 1/(L R_{\max})$, then for any horizon $T$,
\begin{equation}
\label{eq:adam-style-convergence-weighted}
\frac{1}{T}\sum_{t=0}^{T-1}
\mathbb{E}\!\left[
  \langle \nabla F(\theta_t),A_t\nabla F(\theta_t)\rangle
\right]
\leq
\frac{2(F(\theta_0)-F_\star)}{\eta T}.
\end{equation}
Consequently,
\begin{equation}
\label{eq:adam-style-convergence-euclidean}
\frac{1}{T}\sum_{t=0}^{T-1}
\mathbb{E}\!\left[\lVert\nabla F(\theta_t)\rVert^2\right]
\leq
\frac{2(F(\theta_0)-F_\star)}{\eta a_{\min}T}.
\end{equation}

\paragraph{Predictable-preconditioner assumptions and weighted moment.}
The theorem applies to a bounded diagonal preconditioner fixed before the
current tangent draw, for example a deliberately lagged inverse-RMS
preconditioner. It does \emph{not} capture the exact Adam implementation in our
experiments: both its first moment and its second-moment denominator are updated
from the current stochastic estimate before the parameter step. Establishing a
result for that optimizer requires a separate dependence analysis. Within the
idealised algorithm, the trunk-specific step scale $\rho$ of
Appendix~\ref{sec:frozen-control} multiplies the trunk entries of $A_t$ by a
constant and therefore preserves the theorem's assumptions.

For a fixed diagonal preconditioner $A_t$, Split-FG has the exact conditional
weighted second moment
\begin{align}
\label{eq:adam-weighted-second-moment}
\mathbb{E}_t\!\left[\lVert A_t\hat g_t\rVert^2\right]
&=
\lVert A_t g_{\mathrm{head},t}\rVert^2
+\sum_{i\in\mathrm{trunk}} a_{t,i}^2
\left[
  \left(1+\frac{1}{K}\right)g_{t,i}^2
  +\frac{1}{K}\lVert g_{\mathrm{trunk},t}\rVert^2
\right].
\end{align}
This identity explains the possible role of predictable inverse-RMS scaling:
noisy trunk coordinates with large past running second moment can acquire
smaller $a_{t,i}$, so
their contribution to the smoothness penalty is reduced quadratically, while
the exact head coordinates carry no forward-gradient variance term. A coarse
worst-case bound is
$R_{\max}\leq (a_{\max}^2/a_{\min})(1+(P_{\mathrm{trunk}}+1)/K)$, recovering
the unpreconditioned Split-FG scaling up to conditioning of the diagonal
preconditioner.

If an implementation shares a parameter between the trunk computation and the
exact head, the theorem applies to the strict estimator that includes both
contributions (the \texttt{include\_tied} variant). It does \emph{not} cover
the readout-only surrogate used in our GPT runs: there the stop-gradient copy
of the lookup role is redefined at every iterate, so no fixed $L$-smooth,
lower-bounded objective telescopes in the descent argument; the guarantee
applies to the untied parameters, and the tied table's update is a heuristic
we disclose (Appendix~\ref{app:gpt-arch}).

\paragraph{Proof of the predictable-preconditioner theorem.}
For a trunk coordinate $i$, Eq.~\eqref{eq:split-coordinate-var} gives
\begin{equation}
\mathbb{E}_t[\hat g_{t,i}^2]
=
\left(1+\frac{1}{K}\right)g_{t,i}^2
+\frac{1}{K}\lVert g_{\mathrm{trunk},t}\rVert^2 ,
\end{equation}
where $\mathbb{E}_t[\cdot]$ conditions on the current iterate and all previous
randomness. For head coordinates, Split-FG is exact:
$\hat g_{t,i}=g_{t,i}$. Since $A_t$ is fixed under $\mathbb{E}_t[\cdot]$,
multiplying by $a_{t,i}^2$ and summing over coordinates proves the weighted
second-moment identity in Eq.~\eqref{eq:adam-weighted-second-moment}.

By $L$-smoothness and the update
$\theta_{t+1}=\theta_t-\eta A_t\hat g_t$,
\begin{align}
\mathbb{E}_t[F(\theta_{t+1})]
&\leq
F(\theta_t)
-\eta
\left\langle
  \nabla F(\theta_t),A_t\mathbb{E}_t[\hat g_t]
\right\rangle
+\frac{L\eta^2}{2}
\mathbb{E}_t[\lVert A_t\hat g_t\rVert^2]\\
&=
F(\theta_t)
-\eta
\langle \nabla F(\theta_t),A_t\nabla F(\theta_t)\rangle
+\frac{L\eta^2}{2}
\mathbb{E}_t[\lVert A_t\hat g_t\rVert^2]\\
&\leq
F(\theta_t)
-\eta
\left(1-\frac{L\eta R_{\max}}{2}\right)
\langle \nabla F(\theta_t),A_t\nabla F(\theta_t)\rangle\\
&\leq
F(\theta_t)
-\frac{\eta}{2}
\langle \nabla F(\theta_t),A_t\nabla F(\theta_t)\rangle .
\end{align}
The second line uses Split-FG unbiasedness, and the third uses the definition
of $R_t(A_t)$ together with $R_t(A_t)\leq R_{\max}$. Taking total expectation,
summing from $t=0$ to $T-1$, and using $F(\theta_T)\geq F_\star$ gives
Eq.~\eqref{eq:adam-style-convergence-weighted}. Finally,
$A_t\succeq a_{\min}I$ implies
$\langle\nabla F(\theta_t),A_t\nabla F(\theta_t)\rangle
\geq a_{\min}\lVert\nabla F(\theta_t)\rVert^2$, yielding
Eq.~\eqref{eq:adam-style-convergence-euclidean}.

\paragraph{SGD baseline theorem.}
For comparison, the unpreconditioned Split-FG update also converges under the
standard nonconvex stochastic-gradient argument. Assume $F$ is differentiable,
bounded below by $F_\star$, and $L$-smooth. Run the SGD version of Split-FG,
\begin{equation}
\theta_{t+1}=\theta_t-\eta\hat g_t,
\end{equation}
where the head gradient is exact and the trunk estimator is
Eq.~\eqref{eq:split-estimator-analysis}. If
\begin{equation}
\label{eq:split-sgd-stepsize}
0<\eta\leq
\frac{1}{
L\left(1+(P_{\mathrm{trunk}}+1)/K\right)},
\end{equation}
then for any horizon $T$,
\begin{equation}
\label{eq:split-sgd-convergence}
\frac{1}{T}\sum_{t=0}^{T-1}
\mathbb{E}\!\left[\lVert\nabla F(\theta_t)\rVert^2\right]
\leq
\frac{2\left(F(\theta_0)-F_\star\right)}{\eta T}.
\end{equation}
Taking the largest allowed step size yields
\begin{equation}
\min_{0\leq t<T}
\mathbb{E}\!\left[\lVert\nabla F(\theta_t)\rVert^2\right]
\leq
\frac{
2L\left(1+(P_{\mathrm{trunk}}+1)/K\right)
\left(F(\theta_0)-F_\star\right)
}{T}.
\end{equation}
Thus full-batch Split-FG reaches a first-order stationary point at the standard
nonconvex stochastic-gradient rate, with the forward-gradient penalty depending
on $P_{\mathrm{trunk}}$ rather than $P_{\mathrm{total}}$.

\paragraph{Proof of the SGD baseline.}
By $L$-smoothness and the update rule,
\begin{align}
\mathbb{E}_t[F(\theta_{t+1})]
&\leq
F(\theta_t)
-\eta
\left\langle
  \nabla F(\theta_t),\mathbb{E}_t[\hat g_t]
\right\rangle
+\frac{L\eta^2}{2}\mathbb{E}_t[\lVert\hat g_t\rVert^2]\\
&=
F(\theta_t)
-\eta\lVert\nabla F(\theta_t)\rVert^2
+\frac{L\eta^2}{2}
\left[
  \lVert g_{\mathrm{head},t}\rVert^2
  +
  \left(1+\frac{P_{\mathrm{trunk}}+1}{K}\right)
  \lVert g_{\mathrm{trunk},t}\rVert^2
\right]\\
&\leq
F(\theta_t)
-\eta
\left[
  1-\frac{L\eta}{2}
  \left(1+\frac{P_{\mathrm{trunk}}+1}{K}\right)
\right]
\lVert\nabla F(\theta_t)\rVert^2\\
&\leq
F(\theta_t)
-\frac{\eta}{2}
\lVert\nabla F(\theta_t)\rVert^2 .
\end{align}
The second line uses unbiasedness and the second-moment identity from
Eq.~\eqref{eq:split-total-var}; the final line uses
Eq.~\eqref{eq:split-sgd-stepsize}. Taking total expectation, summing over
$t=0,\ldots,T-1$, and using $F(\theta_T)\geq F_\star$ proves
Eq.~\eqref{eq:split-sgd-convergence}.

\paragraph{Minibatches.}
With minibatches, the same argument applies after adding the usual data-noise
term. If the combined estimator remains unbiased for $\nabla F(\theta_t)$ and
has conditional second moment at most
$\left(1+(P_{\mathrm{trunk}}+1)/K\right)\lVert\nabla F(\theta_t)\rVert^2
+\sigma_{\mathrm{data}}^2$, then the SGD bound becomes
\begin{equation}
\frac{1}{T}\sum_{t=0}^{T-1}
\mathbb{E}\!\left[\lVert\nabla F(\theta_t)\rVert^2\right]
\leq
\frac{2\left(F(\theta_0)-F_\star\right)}{\eta T}
+L\eta\sigma_{\mathrm{data}}^2.
\end{equation}
The additional term is the standard minibatch stochastic-gradient term; the
structural Split-FG contribution still scales with $P_{\mathrm{trunk}}/K$.


\section{General Experimental Protocol}
\label{app:general-protocol}

\paragraph{Backpropagation-reference protocol.}
The tabular and GPT comparisons use the same dataset, preprocessing,
architecture, loss function, data exposure, and optimizer family for Split-FG
and their backpropagation references. For language models this means the same
number of training tokens and optimizer updates. The image headline table is a
system comparison: Split-FG uses a light-trunk/flat-head architecture while the
backpropagation, Pure-FG, and ES references use standard full ResNets and the
other baselines use their native architectures. The image appendix therefore
reports the architecture-matched flat-head estimator ablation separately.
Wall-clock time is reported as an outcome, not a matching criterion.

\paragraph{Memory metric.}
For each large experiment we report peak GPU memory and the relative difference
of Split-FG from its stated backpropagation reference:
\begin{equation}
\label{eq:memory-saving}
    \mathrm{MemSave}
    = 1 - \frac{
        M_{\mathrm{peak}}(\mathrm{Split\text{-}FG})
      }{
        M_{\mathrm{peak}}(\mathrm{Backprop})
      } .
\end{equation}
We also report held-out-fold, development, or validation performance as labelled
and step time; the ablation tables
additionally report JVPs per step, and every Split-FG run uses zero
reverse-mode passes through the trunk.

\paragraph{Gradient clipping.} Every run in the paper---backpropagation
included---applies global-norm clipping at $1.0$ to the assembled gradient
estimate (all parameter groups jointly) before the optimizer step. For
high-variance estimates whose norm exceeds the threshold, clipping rescales
the whole update and thereby introduces a bias not covered by the analysis of
Section~\ref{sec:theory}, which treats the unclipped estimator; the protocol
is uniform across methods, so comparisons within each table share it.

\paragraph{Backprop-free baselines.}
The \emph{Forward-Forward} algorithm
\citep{hinton2022forwardforwardalgorithmpreliminaryinvestigations} trains each
layer with a local goodness objective on positive and negative (label-overlaid)
inputs and is defined for classification only; it therefore appears in the
tabular-classification and image experiments but not in regression or language
modelling. \emph{Predictive coding}
\citep{Whittington2017AnAO,millidge2022predictive} settles layer-wise latent
variables to minimise local prediction errors and then applies local Hebbian
weight updates; it applies to both classification and regression and, in the
small-step-size limit, approximates the backpropagation gradient using only
local computation. The \emph{modern Hopfield} associative-memory baseline
\citep{ramsauer2021hopfield} performs a training-free softmax retrieval over a
frozen trunk, storing $(\text{feature},\text{target})$ pairs and answering
queries by $w=\mathrm{softmax}(\beta\, q K^\top)$, $\hat{y}=wV$; for language
modelling this reduces to per-token nearest-neighbour recall in the spirit of
kNN-LM \citep{khandelwal2020knnlm}. All three baselines use zero JVPs and zero
trunk backward passes, isolating how much of each task is solvable by local
goodness training, local error settling, or associative recall over a random
representation.

\section{Tabular Data}
\label{datasets}
\subsection{Tabular Dataset Details}
\label{app:tabular-datasets}

Table~\ref{tab:tabular-datasets} summarizes the eight default tabular datasets
used in the experiments of Section~\ref{sec:tabular-experiment}. They are
standard public benchmarks obtained from OpenML (with Kaggle origins for
Otto Group Products and Diamond), chosen to cover binary classification,
multiclass classification, and regression across a range of feature counts and
dataset sizes. Following the preprocessing in our loader, numeric features are
standardised to zero mean and unit variance and categorical features are
integer-encoded. We use $5$-fold cross-validation for all datasets and methods;
each fold trains under the fixed $300$-step budget of
Appendix~\ref{app:tabular-protocol}.

\begin{table}[h]
\caption{Summary of the default tabular datasets used in
Section~\ref{sec:tabular-experiment}. ``\#Feat.'' is the number of input
features and ``\#Cls.'' the number of target classes (``--'' for regression).
.}
\label{tab:tabular-datasets}
\begin{center}
\small
\resizebox{\textwidth}{!}{%
\begin{tabular}{lllrrrl}
\hline
Dataset & Domain & Task & \#Samples & \#Feat. & \#Cls. & Source \\
\hline
Adult               & Census income      & Binary classification     & $48{,}842$  & $14$ & $2$ & OpenML \\
Higgs Small         & Particle physics   & Binary classification     & $98{,}050$  & $28$ & $2$ & OpenML \\
Otto Group Products & E-commerce         & Multiclass classification & $61{,}878$  & $93$ & $9$ & Kaggle/OpenML \\
Covertype           & Forest cartography & Multiclass classification & $581{,}012$ & $54$ & $7$ & OpenML \\
California Housing  & Housing prices     & Regression                & $20{,}640$  & $8$  & -- & OpenML \\
House 16H           & Housing prices     & Regression                & $22{,}784$  & $16$ & -- & OpenML \\
Diamond             & Diamond prices     & Regression                & $53{,}940$  & $9$  & -- & Kaggle/OpenML \\
Black Friday        & Retail purchases   & Regression                & $166{,}821$ & $9$  & -- & OpenML \\
\hline
\end{tabular}}
\end{center}
\end{table}

\subsection{Tabular Experimental Protocol}
\label{app:tabular-protocol}

\paragraph{Metrics.} For the classification datasets we report held-out-fold
accuracy (\%, higher is better); for the regression datasets we report
held-out-fold RMSE (lower is better) on the standardised targets. Because the
loader standardises regression targets using training-fold statistics, RMSE is expressed in units of target
standard deviations. Classification is trained with cross-entropy and regression
with MSE.

\paragraph{Methods.} Each dataset is run with matched backpropagation, pure
forward gradient (Pure FG), antithetic evolution strategies (ES), and the
no-global-backward baselines Forward-Forward (FF), predictive coding (PC), and the
Hopfield associative-memory readout (Hop.), against Split-FG. FF applies only to
classification (``--'' for regression); PC and Hopfield apply to both. Every
Split-FG run is paired with the matched backpropagation version of the same
TabM-style model (Appendix~\ref{app:tabular-arch}).

\paragraph{Training budget.} Every method uses a fixed budget of $300$ optimiser
steps at batch size $256$, run once per cross-validation fold with seed $0$;
the $\pm$ values in Table~\ref{tab:tabular-main} are standard deviations over
the $5$ folds, not over seeds. The effective number
of epochs is $\mathrm{steps}\times\mathrm{batch}/N_{\mathrm{train}}$ and so
varies with dataset size (reported in Table~\ref{tab:tabular-memory}): under this
controlled budget the largest datasets (Covertype, Black Friday) see well under
one epoch.

\paragraph{Memory measurement.} Table~\ref{tab:tabular-memory} reports peak GPU
memory via \texttt{torch.cuda.max\_memory\_allocated}, reset at the start of
training, and $\mathrm{MemSave}=1-M_{\mathrm{peak}}(\text{Split-FG})/
M_{\mathrm{peak}}(\text{Backprop})$ (Eq.~\eqref{eq:memory-saving}). Both methods
keep the full dataset resident on-device, so the peak includes the data tensors;
this dilutes the activation-graph saving on the largest datasets, which is why
the relative saving is largest for the smallest dataset (California Housing,
$24.1\%$) and smallest for the largest (Black Friday, $4.3\%$). All runs use a
single NVIDIA RTX~4070 laptop GPU.

\begin{table}[h]
\caption{Tabular memory savings: peak GPU memory (MiB) for matched
backpropagation vs.\ Split-FG, with $\mathrm{MemSave}$ from
Eq.~\eqref{eq:memory-saving}. Measurement setup and the dataset-size trend are
detailed in Appendix~\ref{app:tabular-protocol}.}
\label{tab:tabular-memory}
\begin{center}
\small
\begin{tabular}{lccccc}
\hline
Dataset & Loss & Epochs & Backprop mem.\ (MiB) & Split-FG mem.\ (MiB) & MemSave \\
\hline
Adult              & CE  & $2.3$ & $144.4$ & $126.0$ & $12.7\%$ \\
Covertype          & CE  & $0.2$ & $309.8$ & $291.1$ & $6.0\%$ \\
California Housing & MSE & $5.3$ & $76.0$  & $57.7$  & $24.1\%$ \\
Diamond            & MSE & $2.0$ & $156.9$ & $138.6$ & $11.7\%$ \\
Black Friday       & MSE & $0.7$ & $425.2$ & $406.9$ & $4.3\%$ \\
\hline
\end{tabular}
\end{center}
\end{table}

\subsection{Tabular Model Architecture}
\label{app:tabular-arch}

All tabular experiments (Table~\ref{tab:tabular-main}) use the same TabM-style parameter-efficient MLP
ensemble \citep{gorishniy2025tabm}. We reserve $K$ for the number of
forward-gradient tangent samples and write the TabM ensemble count as
$k_{\mathrm{ens}}$. A single shared MLP is evaluated by $k_{\mathrm{ens}}$
ensemble members that differ only through BatchEnsemble rank-1 multiplicative
adapters: at each layer member $i$ computes
$y_i = \big((x_i \odot r_i)\,W^\top\big)\odot s_i + b$, with a shared weight $W$
and bias $b$ and per-member vectors $r_i, s_i$ (initialised near $1$ to break
symmetry). The per-member representations are averaged before a single shared
linear head; because averaging is linear, this keeps the head a single linear
layer, so the trunk/head split admits the closed-form CE/MSE head gradient of
Eq.~\eqref{eq:head-grads}. For the MSE case, with
$\hat{y}=hW_{\mathrm{head}}^\top+b$ and
$\mathcal{L}=(2N)^{-1}\lVert \hat{y}-y\rVert^2$, the closed forms are
$\partial\mathcal{L}/\partial W_{\mathrm{head}} = r^\top h$,
$\partial\mathcal{L}/\partial b=\sum_n r_n$, and
$\partial\mathcal{L}/\partial h=rW_{\mathrm{head}}$, where
$r=(\hat{y}-y)/N$. The trunk (input broadcast, the
$n_{\mathrm{layers}}$ BatchEnsemble layers, and the member average) is the
nonlinear part estimated by forward-gradient JVPs; the linear head is estimated
exactly. Table~\ref{tab:tabular-arch} lists the structure.

Default hyperparameters: hidden width $d_h=128$, $n_{\mathrm{layers}}=2$ shared
layers, $k_{\mathrm{ens}}=8$ ensemble members, $K=8$ forward-gradient tangent samples, the Adam
optimizer at learning rate $3\times10^{-3}$, batch size $256$, trained for $300$
steps. Here $n$ denotes the number of input features and $C$ the number of
outputs ($C$ classes for classification, $C=1$ for regression), both
dataset-dependent (Table~\ref{tab:tabular-datasets}); $B$ is the batch size.

\begin{table}[t]
\caption{Network structure of the TabM-style model used for the tabular
experiments (Table~\ref{tab:tabular-main}). Rows
above the rule form the trunk, estimated by forward-gradient JVPs; the linear
head below the rule is estimated with the exact analytical head gradient
(Eq.~\eqref{eq:head-grads}). $n$: input features; $d_h=128$: hidden width;
$k_{\mathrm{ens}}=8$: ensemble members; $K$: forward-gradient tangent samples;
$C$: number of outputs; $B$: batch size.}
\label{tab:tabular-arch}
\begin{center}
\small
\resizebox{\textwidth}{!}{%
\begin{tabular}{lllp{5.4cm}}
\hline
Layer & Operation & Output shape & Description \\
\hline
Input     & Standardised features         & $(B, n)$    & Numerical features standardised; categoricals integer-encoded and appended. \\
Broadcast & Replicate to $k_{\mathrm{ens}}$ members      & $(B, k_{\mathrm{ens}}, n)$ & Shared input copied to all $k_{\mathrm{ens}}$ ensemble members. \\
1         & BatchEnsemble Linear $+$ ReLU & $(B, k_{\mathrm{ens}}, d_h)$ & Shared $W_1\in\mathbb{R}^{d_h\times n}$, bias; per-member rank-1 scales $r_1,s_1$. \\
2         & BatchEnsemble Linear $+$ ReLU & $(B, k_{\mathrm{ens}}, d_h)$ & Shared $W_2\in\mathbb{R}^{d_h\times d_h}$, bias; per-member rank-1 scales $r_2,s_2$. \\
Mean      & Average over members          & $(B, d_h)$  & $\bar h=\tfrac{1}{k_{\mathrm{ens}}}\sum_{i=1}^{k_{\mathrm{ens}}} h_i$; linearises the head. \\
\hline
\multicolumn{4}{c}{\textit{--- trunk / head split ---}} \\
\hline
Head      & Linear                        & $(B, C)$    & Shared $W_{\mathrm{head}}\in\mathbb{R}^{C\times d_h}$, bias; averaged logits (cls.) / prediction (reg.). \\
Output    & CE / MSE loss                 & scalar      & Cross-entropy (classification) or $\tfrac{1}{2}$\,MSE (regression). \\
\hline
\end{tabular}}
\end{center}
\end{table}

\section{Image Dataset Details}
\label{app:image-datasets}

Table~\ref{tab:image-datasets} summarizes the image-classification benchmarks
used in Section~\ref{sec:image-experiment}. CIFAR-10 and CIFAR-100
\citep{Krizhevsky2009LearningML} are standard $32\times32$ natural-image datasets with a fixed
$50{,}000$/$10{,}000$ train/test split; Following the preprocessing
in our loader, images are standardised per channel; at training time CIFAR uses
standard augmentation (reflection padding with a random $32\times32$ crop and a
random horizontal flip). Unlike the smaller tabular datasets, which use $5$-fold
cross-validation, the current image artifacts use the fixed official train/test
split. Because the test split was inspected during method development and
evaluation samples random batches from it, we label these values development
evaluation rather than final held-out test results
(Appendix~\ref{app:image-protocol}).

\begin{table}[h]
\caption{Summary of the image-classification datasets used in
Section~\ref{sec:image-experiment}. ``Res.'' is the input resolution and
``\#Cls.'' the number of target classes.}
\label{tab:image-datasets}
\begin{center}
\small
\resizebox{\textwidth}{!}{%
\begin{tabular}{lllrrrl}
\hline
Dataset & Domain & Task & \#Samples & Res. & \#Cls. & Source \\
\hline
CIFAR-10     & Natural images & Multiclass classification & $60{,}000$      & $32\times32$   & $10$   & \citep{Krizhevsky2009LearningML} \\
CIFAR-100    & Natural images & Multiclass classification & $60{,}000$      & $32\times32$   & $100$  & \citep{Krizhevsky2009LearningML}  \\
\hline
\end{tabular}}
\end{center}
\end{table}

\subsection{Image Experimental Protocol}
\label{app:image-protocol}

\paragraph{Metrics.} The backpropagation, Pure-FG, ES, and Split-FG artifacts
report top-1 accuracy (\%, higher is better) by sampling $50$ random batches of
size $128$ from the official CIFAR test split; the FF, PC, and Hopfield runners
evaluate all $10{,}000$ examples. Thus the former are Monte Carlo estimates,
not deterministic full-test passes. Moreover, that split was inspected during
method development, so Table~\ref{tab:image-main} is labelled development
evaluation. All trainable models use cross-entropy. CIFAR-10 and CIFAR-100
entries are reported as mean $\pm$ standard deviation over five seeds.

\paragraph{Methods.} Each dataset is run with a standard-backbone
backpropagation reference, pure
forward gradient (Pure FG), antithetic evolution strategies (ES), and the
no-global-backward baselines Forward-Forward (FF), predictive coding (PC), and the
Hopfield associative-memory readout (Hop.), against Split-FG. Pure FG and ES use
the standard ResNet-20 with \emph{all} parameters estimated: Pure FG by a single
full-network JVP (variance $\propto P_{\mathrm{total}}$) and ES by a two-point
zeroth-order finite difference. FF and PC are self-contained
two-hidden-layer MLPs on the flattened image trained without any global backward
pass. More specifically, FF greedily optimises a layer-local ``goodness'' objective with the
label overlaid on the input, and PC settles per-layer latents and updates weights
by a local Hebbian rule. Hopfield is training-free: a frozen random trunk maps
inputs to features and a modern-Hopfield/$k$NN memory answers queries by softmax
retrieval over stored (feature, target) pairs. Split-FG uses the
heavy-head/light-trunk flat-head restructuring (Appendix~\ref{app:image-arch}).
The standard-ResNet backpropagation row is a same-data-exposure reference, not
an architecture-matched estimator control; the latter is reported in
Appendix~\ref{app:image-ablation}.

\paragraph{Training budget.} CIFAR-10 uses $23{,}437$ steps at batch size $128$
($\approx60$ epochs) for every reported method. CIFAR-100 uses that budget only
for backpropagation and Split-FG; Pure FG uses $2{,}000$ steps ($5.12$ epochs),
ES $400$ ($1.02$ epochs), FF and PC $3{,}000$ ($7.68$ epochs), and Hopfield is
training-free. These early-stopped rows diagnose failure at their stated
budgets but are not comparable as a matched-budget ranking. Trainable image
methods use Adam at learning rate $1\times10^{-3}$ and standard augmentation;
the 60-epoch runs use a cosine schedule. CIFAR-10 uses $200$ warmup steps and no
weight decay; CIFAR-100 uses $300$ warmup steps and weight decay
$5\times10^{-4}$ where applicable (Appendix~\ref{app:image-arch}). The
forward-gradient methods use $K=4$ tangent samples. The $15.4$-epoch
($6000$-step) protocol is retained only for the flat-head estimator ablation in
Appendix~\ref{app:image-ablation}.

\paragraph{Memory measurement.} Table~\ref{tab:image-memory} reports peak GPU
memory via \texttt{torch.cuda.max\_memory\_allocated}, reset at the start of
training, and $\mathrm{MemSave}=1-M_{\mathrm{peak}}(\text{Split-FG})/
M_{\mathrm{peak}}(\text{Backprop})$ (Eq.~\eqref{eq:memory-saving}). Unlike the
small tabular and linear models, a convolutional trunk carries large activation
maps, and forward mode carries a tangent copy of the trunk activations; because
the $K$ tangents are evaluated sequentially, peak memory does not grow with $K$.
The comparison therefore spans architectures: the flat-head Split-FG model
versus the full ResNet-20 backprop reference. For reference, carrying
\emph{full-network} tangents on the standard CIFAR-10 backbone (pure FG at
$K{=}4$, memory-identical to the naive full-trunk split) peaks at
$2668$\,MiB---the blow-up quoted in Section~\ref{sec:image-experiment}.


\begin{table}[h]
\caption{Image system peak GPU memory (MiB), measured with
\texttt{torch.cuda.max\_memory\_allocated}; peak memory is per-step and
independent of epoch budget. Backpropagation uses the full standard backbone;
Split-FG uses the light-trunk/flat-head architecture. ``RefDiff'' is their
fractional peak-memory difference and must not be attributed to the estimator
alone. ``Top-1 gap'' is the corresponding cross-system difference after $60$
epochs (mean $\pm$ SD over five seeds).}
\label{tab:image-memory}
\begin{center}
\small
\resizebox{\textwidth}{!}{%
\begin{tabular}{llccccc}
\hline
Dataset & BP backbone & Epochs & Backprop mem.\ (MiB) & Split-FG mem.\ (MiB) & RefDiff & Top-1 gap \\
\hline
CIFAR-10     & ResNet-20 & $60$ & $1520$ & $1282$ & $+15.6\%$ & $-27.7$ \\
CIFAR-100    & ResNet-18 & $60$   & $3402$ & $3117$ & $+8.4\%$  & $-30.4_{\pm1.4}$ \\
\hline
\end{tabular}}
\end{center}
\end{table}

\subsection{Image Model Architecture}
\label{app:image-arch}

\paragraph{Backprop, Pure FG and ES backbones.} CIFAR-10 uses a standard CIFAR
ResNet-20 \citep{he2016deep}: a $3\times3$ stem ($3\to16$) followed by three
stages of three BasicBlocks at widths $16/32/64$ (spatial $32/16/8$), global
average pooling and a linear head---$\approx\!272$k parameters. The harder
$100$-class CIFAR-100 uses the larger ResNet-18: a $3\times3$ stem ($3\to64$, the
CIFAR adaptation with no max-pool) and four stages of two BasicBlocks at widths
$64/128/256/512$ (spatial $32/16/8/4$), global average pooling and a linear
head---$\approx\!11.2$M parameters. Both replace BatchNorm with GroupNorm, so the
trunk is a pure function of its parameters and is differentiable under
\texttt{torch.func.jvp} in \texttt{.eval()} mode.

\paragraph{Split-FG flat head.} Split-FG keeps only the stem and the
\emph{first} BasicBlock as the forward-mode trunk and flattens its feature map
into a large two-layer \emph{linear} head $W_2 W_1$ trained by the exact
closed-form gradient (no activation, no backward pass through the trunk;
Table~\ref{tab:image-arch}). The first block inherits the backbone's first-stage
width, so the trunk is $5{,}136$ parameters (flattened feature
$16\times32\times32{=}16{,}384$, $H{=}64$) for the ResNet-20 on CIFAR-10, and
$75{,}840$ parameters (flattened $64\times32\times32{=}65{,}536$, $H{=}128$) for
the ResNet-18 on CIFAR-100. In both, the trunk is under $1\%$ of the head's
parameters, and $H$ exceeds the class count so the factored head stays full rank.

\paragraph{FF and PC.} Forward-Forward and predictive coding are MLP-based
local-learning rules and cannot use a convolutional backbone; we run them on a
parameter-matched MLP over the flattened image ($3072\to2000\to2000$,
$\approx\!10$M parameters, comparable to the ResNet-18).

\paragraph{Hyperparameters.} GroupNorm with $8$ groups, the Adam optimizer, and
$K{=}4$ forward-gradient tangents throughout. CIFAR-10: learning rate
$1\times10^{-3}$, cosine schedule ($200$ warmup), batch $128$, $60$ epochs (the
$6000$-step / $15.4$-epoch budget is used only for the ablation of
Appendix~\ref{app:image-ablation}). CIFAR-100 backpropagation and Split-FG:
learning rate $1\times10^{-3}$, weight decay $5\times10^{-4}$, cosine schedule
($300$ warmup), batch $128$, $60$ epochs; the other methods use the shorter
budgets stated in Appendix~\ref{app:image-protocol}. Here
$C$ is the number of classes and $B$ the batch size.

\begin{table}[h]
\caption{Network structure of the flat-head ResNet used by Split-FG. Rows above
the rule form the forward-mode trunk (stem $+$ one BasicBlock at the backbone's
first-stage width $w$), estimated by forward-gradient JVPs; the two-layer linear
head below the rule is estimated by the exact closed-form gradient. Concrete
sizes: $w{=}16$, flatten $16{,}384$, $H{=}64$ for the ResNet-20 (CIFAR-10);
$w{=}64$, flatten $65{,}536$, $H{=}128$ for the ResNet-18 (CIFAR-100). $C$:
classes; $B$: batch. Backprop and the FG/ES baselines instead continue the full
backbone body (remaining stages, global pool, linear head) in place of the flat
head.}
\label{tab:image-arch}
\begin{center}
\small
\resizebox{\textwidth}{!}{%
\begin{tabular}{lllp{5.4cm}}
\hline
Layer & Operation & Output shape & Description \\
\hline
Input    & Image                     & $(B,3,32,32)$  & Per-channel standardised; train-time reflect-crop $+$ horizontal flip. \\
Stem     & $3\times3$ Conv, GN, ReLU & $(B,w,32,32)$  & $3\to w$ channels, stride $1$, padding $1$; GroupNorm ($8$ groups). \\
Block 1  & BasicBlock                & $(B,w,32,32)$  & Two $3\times3$ convs $w\to w$ (Conv-GN-ReLU, Conv-GN) with identity shortcut, then ReLU. \\
Flatten  & Reshape                   & $(B,1024w)$    & The $w\times32\times32$ feature map flattened ($1024w$). \\
\hline
\multicolumn{4}{c}{\textit{--- trunk / head split ---}} \\
\hline
Head 1   & Linear                    & $(B, H)$       & Shared $W_1\in\mathbb{R}^{H\times 1024w}$, bias; no activation. \\
Head 2   & Linear                    & $(B, C)$       & Shared $W_2\in\mathbb{R}^{C\times H}$, bias; logits $=W_2(W_1 h)+b$ (factored linear). \\
Output   & CE loss                   & scalar         & Cross-entropy over $C$ classes. \\
\hline
\end{tabular}}
\end{center}
\end{table}

\subsection{Flat-head Estimator Ablation}
\label{app:image-ablation}

To separate the contribution of Split-FG's exact head from the choice of
architecture, Table~\ref{tab:image-ablation} runs every estimator on the
\emph{same} flat-head network (CIFAR-10, the $15.4$-epoch/$6000$-step budget,
which is why its Split-FG entry sits below the $60$-epoch $60.5\%$ of
Table~\ref{tab:image-main}). Pure FG and ES estimate \emph{all}
$\approx\!1.05$M parameters --- the two-layer linear head alone is $99.5\%$ of
them --- whereas Split-FG estimates only the $5{,}136$-parameter trunk and
computes the head gradient exactly. Treating the head exactly is worth $+34$
points over forward-estimating it ($56.9$ vs $22.6$): on a fixed network, the
variance of estimating a million-parameter head ($\propto P_{\mathrm{total}}$)
is precisely what the split removes, while backprop on the same net ($70.7\%$)
bounds the architecture. The table also includes Split-FG with a \emph{single}
exact linear head ($54.4_{\pm1.5}$): the factored two-layer head of
Section~\ref{sec:image-experiment} improves on it by $+2.5$ points (paired
$t(4)=6.93$, $p<0.01$), the deep-linear over-parameterisation effect cited in
the main text. This is the architecture-matched comparison
underlying the Split-FG column of Table~\ref{tab:image-main}.

\begin{table}[h]
\caption{Flat-head estimator ablation (CIFAR-10, $15.4$-epoch/$6000$-step
budget): all estimators on the identical
flat-head network (one-block forward-mode trunk $+$ two-layer linear head). Pure
FG and ES estimate every parameter; Split-FG estimates only the trunk and treats
the head exactly. Top-1 accuracy (\%), mean $\pm$ std ($2$ seeds for
backprop/Pure FG/ES, $5$ for the Split-FG variants).}
\label{tab:image-ablation}
\begin{center}
\small
\begin{tabular}{llc}
\hline
Estimator on the flat-head net & Params estimated & Top-1 (\%) \\
\hline
Backprop (exact reverse mode)     & --- (exact)             & $70.7_{\pm0.3}$ \\
\textbf{Split-FG} (exact $2$-layer linear head) & trunk only ($5{,}136$)  & $\mathbf{56.9_{\pm1.2}}$ \\
Split-FG (exact single linear head) & trunk only ($5{,}136$) & $54.4_{\pm1.5}$ \\
Pure FG (forward gradient)        & all ($1.05$M)           & $22.6_{\pm1.4}$ \\
ES (antithetic, zeroth-order)     & all ($1.05$M)           & $21.7_{\pm1.9}$ \\
\hline
\end{tabular}
\end{center}
\end{table}

\section{GPT-Style Language-Modelling Details}
\label{app:gpt-datasets}

Table~\ref{tab:gpt-datasets} summarizes the language-modelling dataset used in
Section~\ref{sec:main-experiment}. We use WikiText-103 raw
\citep{merity2017pointer}, tokenized with the GPT-2 byte-pair encoding \citep{radford2019language}
(\(50{,}257\) tokens). In summary, we have \(119{,}085{,}169\) train tokens and
\(249{,}750\) validation tokens.

\begin{table}[h]
\caption{Summary of the GPT-style language-modelling dataset. Token counts are
after GPT-2 BPE tokenization and insertion of end-of-text separators. The full
validation pass uses complete non-overlapping \(T{=}128\) windows, covering
\(249{,}728\) target tokens.}
\label{tab:gpt-datasets}
\begin{center}
\small
\begin{tabular}{lllrrl}
\hline
Dataset & Split & Tokenizer & Tokens & Eval tokens & Source \\
\hline
WikiText-103 & Train & GPT-2 BPE & \(119{,}085{,}169\) & -- & \texttt{wikitext-103-raw-v1} \\
WikiText-103 & Val.  & GPT-2 BPE & \(249{,}750\) & \(249{,}728\) & \texttt{wikitext-103-raw-v1} \\
\hline
\end{tabular}
\end{center}
\end{table}

\subsection{GPT Experimental Protocol}
\label{app:gpt-protocol}

\paragraph{Metrics.} We report validation negative log-likelihood (NLL) and
perplexity, where \(\mathrm{PPL}=\exp(\mathrm{NLL})\), together with wall-clock
time, mean step time, peak GPU memory, number of JVPs per step, and backward
trunk passes. Validation is a full pass over the validation token stream using
non-overlapping windows; no validation minibatches are sampled for the reported
main run.

\paragraph{Methods.} The main GPT comparison uses same-architecture,
same-token-exposure backpropagation with
Adam and no weight decay (the recipe used by Split-FG; an AdamW variant with
decoupled weight decay $0.01$ reaches perplexity $75.5$ and is noted in the
caption of Table~\ref{tab:main-lm}), pure forward gradient with Adam, pure
forward gradient with SGD,
antithetic ES with Adam, the Hopfield associative-memory/kNN-LM baseline, and
Split-FG with Adam. Forward-Forward is omitted because it is not defined for
next-token language modelling; predictive coding and equilibrium propagation
are omitted because our implementations of those baselines do not extend to
next-token language modelling. The Hopfield/kNN-LM row is
training-free: a frozen random GPT trunk maps training windows to features and
validation queries retrieve next-token distributions by softmax \(k\)NN over the
stored features. On the 8GB run we cap the store at \(4096\) train windows and
evaluate all validation windows; its reported time entry in
Table~\ref{tab:main-lm} is one full retrieval-evaluation pass rather than a
training-step time.

\paragraph{Training budget.} All GPT runs use the \texttt{small8}
GPT configuration in Table~\ref{tab:gpt-arch}: \(L=4\) layers,
\(d_{\mathrm{model}}=256\), \(4\) attention heads,
\(d_{\mathrm{ff}}=1024\), sequence length \(T=128\), and batch size \(4\).
Training uses deterministic non-overlapping windows over the full tokenized
WikiText-103 train split. This gives \(930{,}352\) train windows and
\(232{,}588\) optimizer steps for one epoch. All trainable methods use the same
number of optimizer steps and the same token exposure. The optimizer schedule is
cosine decay with \(100\) warmup steps. The default learning rate is
\(1.5\times10^{-3}\) for Adam/AdamW methods and \(5\times10^{-4}\) for the SGD
forward-gradient baseline. The main table reports Split-FG at \(K{=}4\) with
the reduced trunk step (\(\rho=0.03\),
Appendix~\ref{sec:frozen-control}) alongside the naive-step run at the same
\(K\); the frozen-trunk control uses \(K{=}0\)
(head-only), and the \(K\)-ablation table reports the sweep with the same
reduced trunk step. Mechanically, the trunk step scale multiplies the Adam
update of every trunk parameter by \(\rho\) while the head parameters keep the
base learning rate; it changes neither the estimator nor the memory or step
time.

\paragraph{Gradient-energy measurement.} The gradient-energy ratio quoted in
Section~\ref{split-fg} is measured at initialisation (seed $0$, the same
initialisation as every \texttt{small8} run) by computing exact
backpropagation gradients on $16$ sequential non-overlapping WikiText-103
training minibatches ($T{=}128$, $B{=}4$) and averaging the per-batch ratio
$\lVert\nabla_{\theta_{\mathrm{trunk}}}\mathcal{L}\rVert^2/\lVert\nabla_{\theta}\mathcal{L}\rVert^2$
under the same \texttt{exclude\_tied} trunk/head partition used for training;
the measurement script is included in the supplementary material. The ratio
evolves over training; we report the initialisation value.

\paragraph{Pure-FG baseline tuning.} The $K{=}4$ pure-FG row of
Table~\ref{tab:main-lm} is tuned by the same screening protocol as $\rho$
(fully annealed runs at the $10\%$-epoch budget of $23{,}258$ steps, one full
epoch at the argmin), applied to a six-point learning-rate grid. Screening
validation perplexities: $3\times10^{-3}$: $21{,}319$;
$1.5\times10^{-3}$: $8{,}026$; $4.5\times10^{-4}$: $4{,}544$ (argmin);
$1.5\times10^{-4}$: $6{,}111$; $4.5\times10^{-5}$: $16{,}510$;
$1.5\times10^{-5}$: $32{,}543$. The full-epoch run at the argmin reaches
$2{,}885$ (the value in Table~\ref{tab:main-lm}). The grid is over the global
learning rate because pure FG has no exact/estimated group split; it spans
$200\times$, comparable to the $100\times$ span of the five-point $\rho$ grid
in Table~\ref{tab:rho-ablation}.

\subsection{Frozen-Trunk Control and the Trunk Step Scale}
\label{sec:frozen-control}

Because Split-FG's head is exact, any comparison against other estimators
leaves one question open: does the forward-gradient \emph{trunk} training
contribute, or is the exact head over random features doing all the work? We
answer it with the cheapest possible control: freeze the trunk at its random
initialisation and train only the head with the exact gradient, under an
identical budget and schedule. Table~\ref{tab:main-lm} in the main text
gathers the outcome, including two count-based references fitted on the
training tokens and evaluated under the same windowed protocol (a unigram
model, $1659$, and an interpolated bigram model, $198$);
Figure~\ref{fig:frozen-control} (main text) shows the trajectories.

\paragraph{Naive trunk training hurts at this scale.}
The frozen control reaches perplexity $667.6$: a random causal transformer is
already a strong feature extractor, and the exact head reads substantial
structure off it (well below the unigram reference). Naive Split-FG, by
contrast, converges at $1733.4$ at $K{=}4$ (the run shown in
Figure~\ref{fig:frozen-control}; even $K{=}8$ only reaches
$1656.7$)---\emph{above} the frozen control and near the unigram level.
Training the trunk with the naive step destroys more than it learns.

\paragraph{Diagnosis: a noise-blind step size.}
The trunk estimate is unbiased, but its per-coordinate signal-to-noise ratio
is ${\sim}1/P_{\mathrm{trunk}}$ (Eq.~\eqref{eq:snr}), and one epoch supplies
only $\mathrm{steps}\times K/P_{\mathrm{trunk}}\approx0.3$ scalar probes per
trunk parameter---the trunk is under-determined within the budget. Adam is
blind to this: its second-moment normalisation drives \emph{every} coordinate
at approximately full step length, whether its gradient estimate is signal or
noise. On this account, the trunk executes a near-full-speed random walk away
from its good random initialisation while the exact head continuously re-fits
a drifting, degrading representation. We emphasise that the controls are
consistent with this mechanism rather than a proof of it (we do not trace
trunk displacement directly), but under it the failure is due to the step
size rather than the estimator.

\paragraph{Fix: reduce the trunk step.}
We multiply the trunk's Adam step by a single factor $\rho\!\ll\!1$ (the head
keeps the base rate): the noise-driven random walk should be suppressed
quadratically, while the signal---accumulated coherently by the first-moment
average---still drives a slow, consistent drift. This is a selected per-group
learning-rate multiplier, not an SNR-derived rule. For transparency about the selection:
$\rho=0.03$ was fixed by a single exploratory run at a $10\%$ budget
\emph{before} the sensitivity grid of Appendix~\ref{app:gpt-rho} existed; the
grid, run afterwards, places the optimum near $\rho=0.1$ at both budgets; the
headline run retains the earlier selected value $\rho=0.03$. All selection used
the validation split. An adaptive rule is left to future work. The same trunk
that hurt now helps in this run: perplexity $386.7$, $42\%$
below the frozen control and $4.5\times$ below the $K$-matched naive run, at
identical memory and step time; Figure~\ref{fig:frozen-control} shows the scaled run
below the frozen control at every checkpoint, passing its \emph{final}
perplexity after about a quarter of the budget. The image controls (single
seed) mark the benign end of the scale: with only $5{,}136$ trunk parameters
on CIFAR-10, the unscaled trunk already helps ($60.5\%$ vs.\ $55.2\%$ frozen),
while CIFAR-100 ($75{,}840$) sits near break-even ($35.2\%$ vs.\ $34.8\%$).
Probe density alone does not explain the pattern---a light-trunk GPT variant
with ample probe coverage merely ties the frozen control
(Section~\ref{discussion})---evidence consistent with step size being an
important variable. The
interpolated bigram reference ($198$) remains $2\times$ below even the fixed
method, marking the clear next target.

\subsection{GPT Core Ablation}
\label{app:gpt-core-ablation}

\begin{table}[h]
\caption{Core ablation on WikiText-103; the relative factor is validation
perplexity against the main method (Split-FG + Adam, $K{=}4$, trunk step
$\times0.03$). The first three rows change exactly one component. The fourth
removes the split at \emph{matched} tangent count and tuning ($K{=}4$,
learning rate tuned by the same screening protocol as $\rho$;
Appendix~\ref{app:gpt-protocol})---the fair
attribution of the split's benefit, $7.5\times$; the remaining rows are the
untuned $K{=}1$ Table~\ref{tab:main-lm} baselines.
Entries from Table~\ref{tab:main-lm} and Table~\ref{tab:k-ablation}.}
\label{tab:core-ablation}
\begin{center}
\small
\begin{tabular}{lccc}
\hline
Configuration & Val. PPL & Relative factor & Interpretation \\
\hline
Split-FG + Adam, $K{=}4$, trunk step $\times0.03$ & $386.7$ & $1.0\times$ & main method \\
Undo step scaling (trunk step $\times1$, same $K{=}4$) & $1733.4$ & $4.5\times$ & noise-blind steps \\
Freeze the trunk (head only, $K{=}0$)             & $667.6$  & $1.7\times$ & trunk training helps \\
Remove split (pure FG + Adam, $K{=}4$, tuned lr)  & $2885.0$  & $7.5\times$ & \emph{matched} split attribution \\
Remove split (pure FG + Adam, $K{=}1$, default)   & $40128.9$ & $104\times$ & untuned baseline \\
Remove split (pure FG + SGD, $K{=}1$)             & $50385.1$ & $130\times$ & at the uniform floor \\
Use ES instead of JVP (+ Adam, $K{=}1$)           & $12103.1$ & $31\times$  & zeroth-order comparison \\
\hline
\end{tabular}
\end{center}
\end{table}

\subsection{GPT Trunk-Step-Scale Ablation}
\label{app:gpt-rho}

The step-scale fix of Appendix~\ref{sec:frozen-control} introduces one
hyperparameter, $\rho$. Table~\ref{tab:rho-ablation} sweeps it over two
orders of magnitude at two budgets: a $10\%$-epoch screening budget and the
full epoch, each run fully annealed and compared against a frozen control
($K{=}0$) at the same budget. The dependence is a smooth, asymmetric basin at
both budgets, degrading gracefully toward the frozen bar as $\rho\!\to\!0$
(the trunk barely moves) and falling off more steeply as $\rho\!\to\!1$ (the
full-budget naive run at $\rho{=}1$ reaches $1733.4$;
Table~\ref{tab:core-ablation}). Every grid point beats the frozen control at
both budgets, so the fix is not sensitive to the exact value of $\rho$. The
optimum is stable across budgets: $\rho=0.1$ minimises perplexity at both the
screening
($541.9$) and full ($365.0$) budgets---so the paper's default $\rho=0.03$,
one grid step below the optimum, is a conservative choice. The grid itself is
shown in Table~\ref{tab:rho-ablation} (main text).

\subsection{GPT Optimizer Ablation}
\label{app:gpt-optim}

\begin{table}[h]
\caption{Optimizer ablation on WikiText-103 (\texttt{small8}, one full epoch,
$K{=}4$, trunk step $\times0.03$, single seed). Adam, AdamW, and Muon all
train the split to $382$--$399$ perplexity; SGD, lacking per-coordinate
normalisation, stalls near the random-token floor ($50{,}257$). AdamW uses
decoupled weight decay $0.01$ scaled by the per-parameter learning rate; Muon
applies Nesterov momentum $+$ Newton--Schulz orthogonalisation to the $2$-D
block weights with an internal Adam for the remaining parameters. Step times
are approximate (shared laptop GPU).}
\label{tab:optim-ablation}
\begin{center}
\small
\begin{tabular}{lcccl}
\hline
Optimizer & Val.\ PPL & Val.\ NLL & Step time & Note \\
\hline
Muon              & $382.3$   & $5.946$  & ${\sim}110$ ms & orthogonalised trunk update \\
Adam              & $386.7$   & $5.958$  & ${\sim}134$ ms & main configuration \\
AdamW             & $398.8$   & $5.988$  & ${\sim}141$ ms & decoupled weight decay \\
SGD               & $44917.9$ & $10.713$ & ${\sim}118$ ms & no adaptive normalisation \\
\hline
\end{tabular}
\end{center}
\end{table}

\subsection{GPT Tangent-Count Ablation}
\label{app:gpt-k-ablation}

\begin{figure}[h]
\begin{center}
\includegraphics[width=\linewidth]{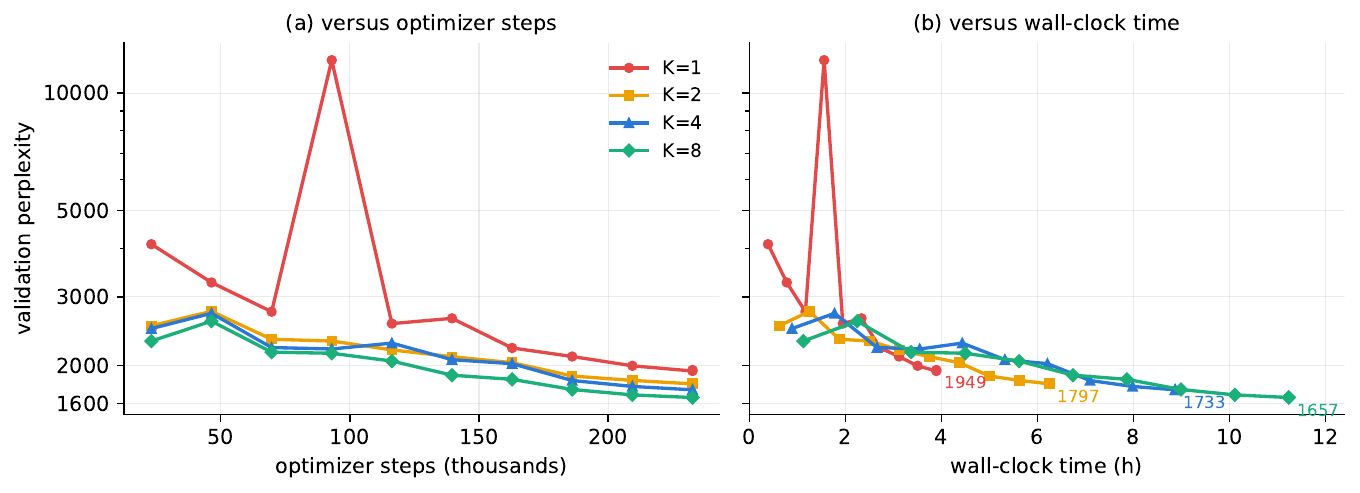}
\end{center}
\caption{Split-FG tangent-count sweep on WikiText-103 with the reduced
trunk step ($\rho=0.03$; \texttt{small8}, one
epoch per run, cosine schedule): validation perplexity versus optimizer steps
(left) and versus wall-clock time (right) for $K\in\{1,2,4,8\}$. Perplexity
improves monotonically in $K$ ($546.8\to361.5$) at constant peak memory
($811$\,MiB; Table~\ref{tab:k-ablation}), and \emph{every} $K$ converges below
the frozen-trunk control (dotted, $667.6$)---even $K{=}1$. On the wall-clock
axis the curves share a common frontier, so $K$ trades time for variance;
comparisons across $K$ are schedule-confounded, since each run anneals over its
own full epoch.}
\label{fig:main-training-curves}
\end{figure}

\begin{table}[h]
\caption{Tangent-count ablation on WikiText-103 with the reduced trunk step
($\rho=0.03$; \texttt{small8}, one full
epoch per run, cosine schedule, single seed; trajectories in
Figure~\ref{fig:main-training-curves}). Increasing $K$ reduces trunk-estimator
variance ($\propto P_{\mathrm{trunk}}/K$): perplexity improves monotonically
with diminishing returns, and \emph{every} $K$---even $K{=}1$---beats the
frozen-trunk control ($667.6$; Table~\ref{tab:main-lm}). Peak
memory is constant because the $K$ tangents are evaluated in a sequential loop
(one JVP tape freed before the next, plus a single gradient accumulator), so
$K$ is a pure time--variance dial; step time is set by the JVP count and is
independent of $\rho$ (measured on a shared laptop GPU).}
\label{tab:k-ablation}
\begin{center}
\begin{tabular}{lccccc}
\hline
$K$ & JVPs/step & Final PPL & Val.\ NLL & Step time & Peak mem. \\
\hline
$1$  & $1$  & $546.8$ & $6.304$ & $52$ ms  & $811$ MiB \\
$2$  & $2$  & $437.6$ & $6.081$ & $62$ ms  & $811$ MiB \\
$4$  & $4$  & $386.7$ & $5.958$ & $134$ ms & $811$ MiB \\
$8$  & $8$  & $361.5$ & $5.890$ & $187$ ms & $811$ MiB \\
\hline
\end{tabular}
\end{center}
\end{table}

\paragraph{Memory measurement.} We report peak GPU memory from
\texttt{torch.cuda.max\_memory\_allocated}, reset at the start of each run. The
same statistic is used for matched backpropagation and Split-FG so that
\(\mathrm{MemSave}\) in Eq.~\eqref{eq:memory-saving} measures the effect of
removing trunk backpropagation. The run is carried out on a single NVIDIA
RTX~4070 laptop GPU with 8GB VRAM. Full validation runs under
\texttt{torch.no\_grad()}, but it is still part of the measured process and is
therefore included in the peak-memory trace if it exceeds the training peak.

\subsection{GPT Model Architecture}
\label{app:gpt-arch}

The GPT model is a causal transformer trained from random initialization. The
default Split-FG decomposition places the split immediately before the linear
LM head: the trunk computation maps token ids to hidden states through the
token lookup, positional embeddings, all transformer blocks, and final
LayerNorm, and the head is a large vocabulary projection
\(hW_{\mathrm{out}}^\top+b\). In the reported runs, the vocabulary table used by
the output projection is assigned to the exact head group, together with the
LM-head bias; its token-lookup role is held fixed with respect to the trunk JVP.
The forward-mode trunk JVP covers the positional table, transformer blocks, and
final LayerNorm.
Table~\ref{tab:gpt-arch} lists the concrete 8GB architecture.

\begin{table}[h]
\caption{Network structure of the 8GB \texttt{small8} GPT model used for the
full WikiText-103 Table~\ref{tab:main-lm} run. Rows above the rule form the
trunk computation for Split-FG. The forward-mode trunk parameters are the
positional table, transformer blocks, and final LayerNorm; the vocabulary table
used by the output projection and the LM-head bias below the rule are trained by
the analytical cross-entropy gradient, while the table's token-lookup role is
held outside the trunk JVP. For this model
\(P_{\mathrm{trunk}}=3{,}192{,}320\),
\(P_{\mathrm{head}}=12{,}916{,}049\), and
\(P_{\mathrm{total}}=16{,}108{,}369\), so
\(P_{\mathrm{trunk}}/P_{\mathrm{total}}=0.198\).}
\label{tab:gpt-arch}
\begin{center}
\small
\resizebox{\linewidth}{!}{%
\begin{tabular}{lllp{5.7cm}}
\hline
Layer & Operation & Output shape & Description \\
\hline
Input & Token ids & \((B,T)\) & Next-token language modelling windows, \(B=4\), \(T=128\). \\
Embedding & Token \(+\) position & \((B,T,256)\) & Token lookup plus learned positional table; the vocabulary table's lookup role is held outside the trunk JVP. \\
Blocks \(1\ldots4\) & Causal self-attention \(+\) MLP & \((B,T,256)\) & Four transformer blocks, \(4\) heads, feed-forward width \(1024\), GELU activations. \\
Final norm & LayerNorm & \((B,T,256)\) & Final normalization before the LM head. \\
\hline
\multicolumn{4}{c}{\textit{--- trunk / head split ---}} \\
\hline
LM head & Linear vocabulary projection & \((B,T,50257)\) & Logits \(=hW_{\mathrm{out}}^\top+b\); the vocabulary table and \(b\) receive the exact CE head gradient. \\
Output & CE loss & scalar & Cross-entropy over all token positions. \\
\hline
\end{tabular}}
\end{center}
\end{table}

\end{document}